\documentclass[conference]{IEEEtran}
\usepackage{times}

\usepackage[numbers]{natbib}
\usepackage{multicol}
\usepackage[bookmarks=true]{hyperref}

\usepackage{bbm}
\usepackage{calc}
\usepackage{url}
\usepackage{hyperref}
\hypersetup{
  colorlinks =false,
  urlcolor = black,
  linkcolor = black
}
\usepackage{graphicx}
\usepackage[cmex10]{amsmath}
\usepackage{bm}
\usepackage{amssymb}
\usepackage{rotating}

\usepackage{chngcntr}
\counterwithin{paragraph}{subsection} 

\usepackage{nicefrac}
\usepackage{cite}
\usepackage[caption=false,font=footnotesize]{subfig}
\usepackage[usenames, dvipsnames]{color}
\usepackage{colortbl}
\usepackage{overpic}
\graphicspath{{./pictures/pdf/},{./pictures/ps/},{./pictures/png/},{./pictures/jpg/}}
\usepackage{breqn} 
\usepackage[ruled]{algorithm}
\usepackage{algpseudocode}
\usepackage{multirow}
\usepackage{todonotes}

\newcommand{\figwid}{0.22\columnwidth}

\allowdisplaybreaks

\begin{document}

\title{Algorithms For Shaping a Particle Swarm With a Shared Control Input Using Boundary Interaction}

\author{Shiva Shahrokhi, Arun Mahadev, and Aaron T. Becker}



%

\newpage

\maketitle

\begin{abstract}
Consider a swarm of particles controlled by global inputs. 
This paper presents algorithms for shaping such swarms in 2D using boundary walls.
The range of configurations created by conforming a swarm to a boundary wall is limited. 
We describe the set of stable configurations of a swarm in two canonical workspaces, a circle and a square. 
To increase the diversity of configurations, we add boundary interaction to our model.  
We provide algorithms using friction with walls to place two robots at arbitrary locations in a rectangular workspace.
Next, we extend this algorithm to place $n$ agents at desired locations. 
We conclude with efficient techniques to control the covariance of a swarm not possible without wall-friction. 
Simulations and hardware implementations with 100 robots validate these results.

These methods may have particular relevance for current micro- and nano-robots controlled by global inputs.

\end{abstract}

\IEEEpeerreviewmaketitle

\section{Introduction}\label{sec:Intro}
Particle swarms steered by a global force are common in applied mathematics, biology, and computer graphics. 
\begin{align}
[\dot{x}_i, \dot{y}_i]^\top = [u_x, u_y]^\top, \qquad i \in [1,n] \label{eq:swarmDynamics}
\end{align}
The control problem is to design $u_x(t), u_y(t)$ to make all $n$ particles achieve a task.
As a current example, micro- and nano-robots can be manufactured in large numbers, see~\citet{Chowdhury2015}, \citet{martel2014computer}, \citet{kim2015imparting}, \citet{Donald2013}, \citet{Ghosh2009}, \citet{Ou2013} or \citet{qiu2015magnetic}.
Someday large swarms of robots will be remotely guided
  ex vivo to assemble structures in parallel and 
 through the human body, to cure disease, heal tissue, and prevent infection. 
 For each application, large numbers of micro robots are required  to deliver sufficient payloads, but the small size of these robots makes it difficult to perform onboard computation.  Instead, these robots are often controlled by a global, broadcast signal. 
These applications require control techniques that can reliably exploit large populations despite high under-actuation.


\begin{figure}
\centering
\begin{overpic}[width=0.95\columnwidth]{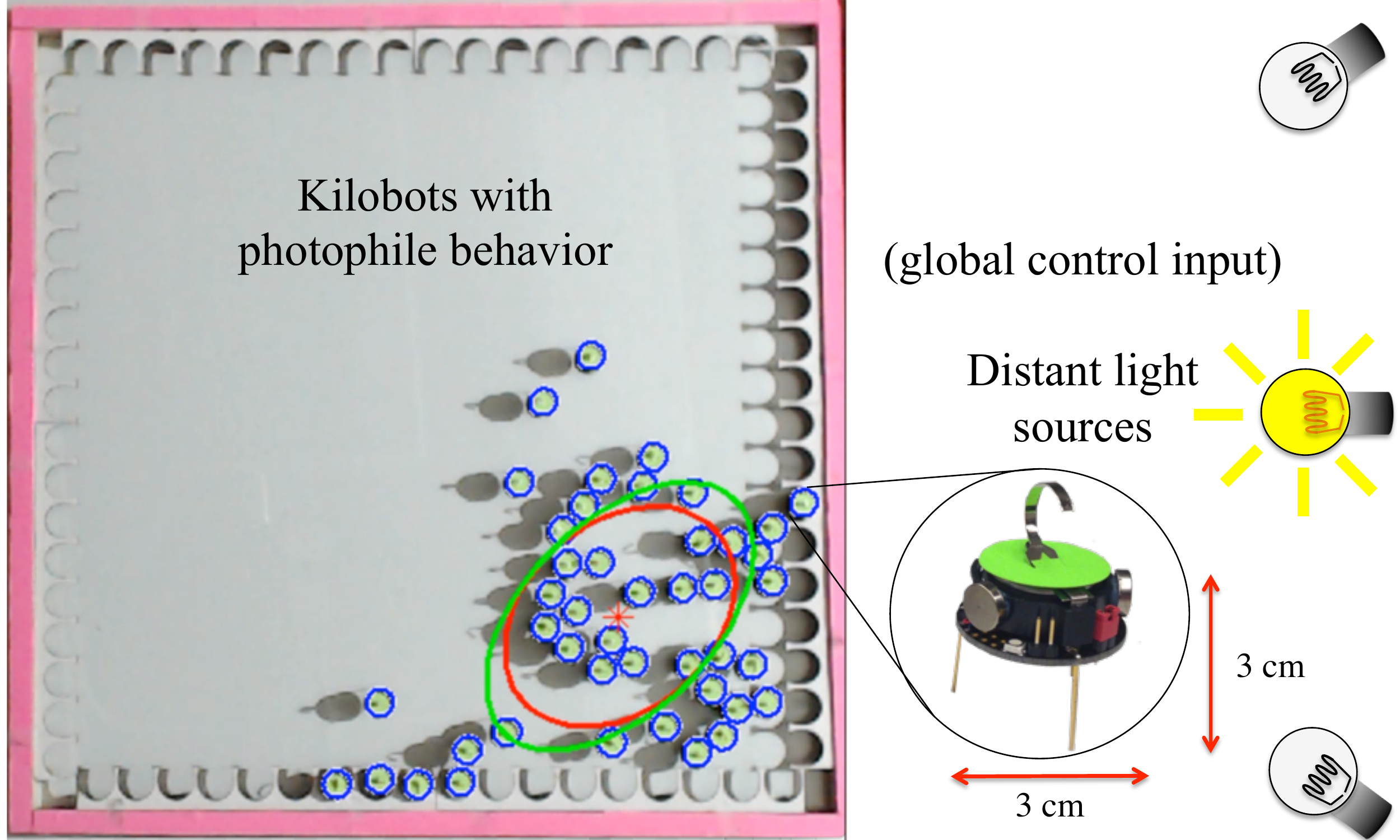}\end{overpic}
\caption{\label{fig:IntroPic}
Swarm of kilobots programmed to move toward the brightest light source as explained in \S \ref{sec:expResults}. The current covariance ellipse and mean are shown in red, the desired covariance is shown in green.  Navigating a swarm using global inputs is challenging because each member receives the same control inputs. 
This paper focuses on using boundary walls and wall friction to break the symmetry caused by the global input and control the shape of a swarm.} 
\end{figure}

Even without obstacles or boundaries, the mean position of the swarm in \eqref{eq:swarmDynamics} is controllable.  By adding rectangular boundary walls, some higher-order moments such as the swarm's position variance orthogonal to the boundary walls ($\sigma_x$ and $\sigma_y$ for a workspace with axis-aligned walls) are also controllable~\citep{ShahrokhiIROS2015}. 
A limitation is that global control can only compress a swarm orthogonal to obstacles.  However, navigating through narrow passages often requires control of the variance and the covariance.

The paper is arranged as follows. \S\ref{subsec:FluidInTank} provides analytical position control results in two canonical workspaces with frictionless walls.  These results are limited in the set of shapes that can be generated.  To extend the range of possible shapes, \S \ref{subsec:WallFriction} introduces wall friction to the system model.  We prove that two orthogonal boundaries with high friction are sufficient to arbitrarily position two robots in \S \ref{sec:PostionControl2Robots}, and \S \ref{sec:PostionControlnRobots} extends this to prove a rectangular workspace with high-friction boundaries can position a swarm of $n$ robots arbitrarily within a subset of the workspace.
\S \ref{sec:simulation} describes implementations of both position control algorithms in simulation and  \S \ref{sec:expResults} describes experiments with a hardware setup and up to 100 robots, as shown in Fig.~\ref{fig:IntroPic}. After a review of recent related work \S \ref{sec:RelatedWork}, we end with directions for future research \S \ref{sec:conclusion}.
\section{Theory}
\label{sec:theory}
\begin{figure*}[!htb]
\begin{center}
\includegraphics[width=\linewidth]{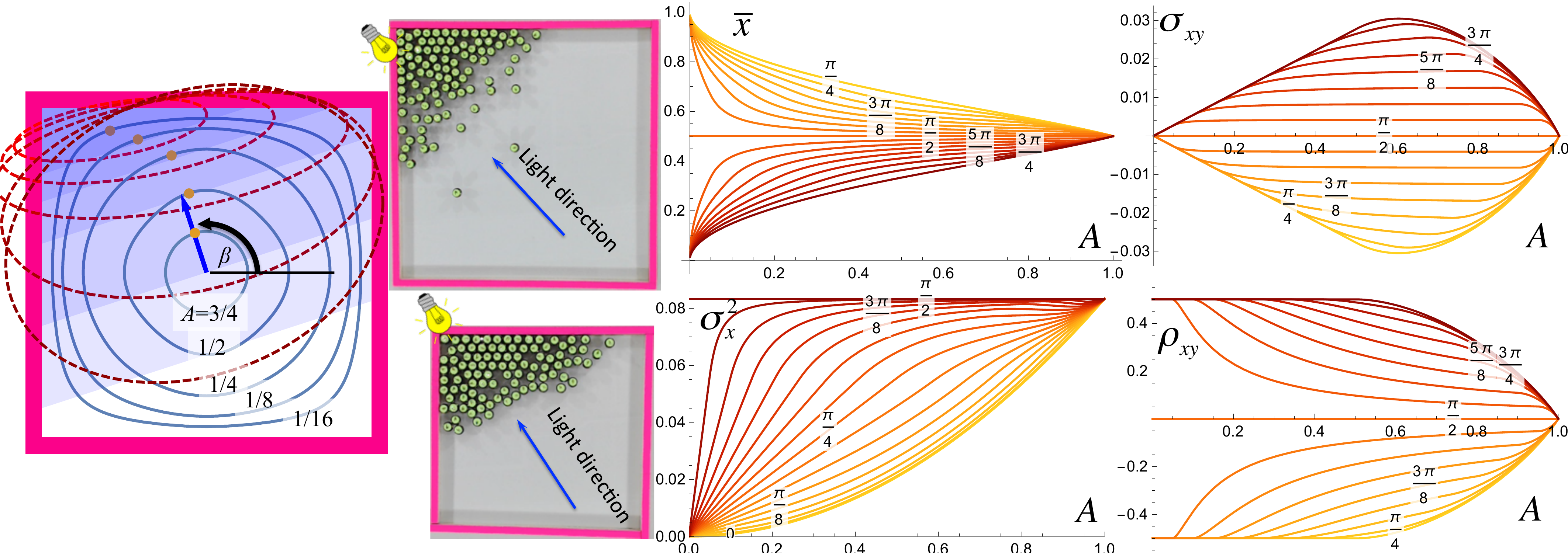} 
\vspace{-1em}
\caption{Pushing the swarm against a square boundary wall allows limited control of the shape of the swarm, as a function of swarm area $A$ and the commanded movement direction $\beta$. Left plot shows locus of possible mean positions for five values of $A$.  The locus morphs from a square to a circle as $A$ increases.  The covariance ellipse for each $A$ is shown with a dashed line. Center shows two corresponding arrangements of kilobots.  At right is $\bar{x}(A), \sigma_{xy}(A), \sigma_x^2(A),$ and $\rho(A)$ for a range of $\beta$ values. See online interactive demonstration at \citep{Zhao2016mathematicaSquare}.}
\label{fig:SquareFill}
\end{center}
\end{figure*} 
\begin{figure*}[!htb]
\begin{center}
\includegraphics[width=\linewidth]{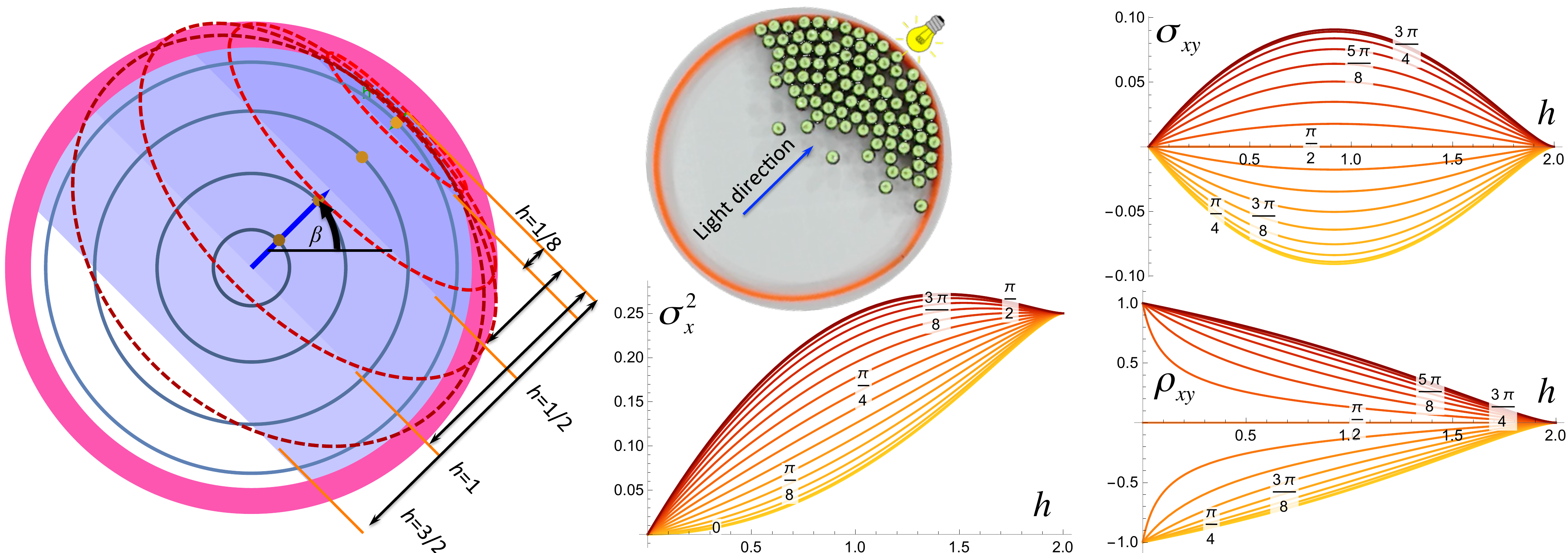} 
\vspace{-1em}
\caption{Pushing the swarm against a circular boundary wall allows limited control of the shape of the swarm, as a function of the fill level $h$ and the commanded movement direction $\beta$. Left plot shows locus of possible mean positions for four values of $h$. The locus of possible mean positions are concentric circles. See online interactive demonstration at \citep{Zhao2016mathematica}.}  
\label{fig:CircleFill}
\end{center}
\end{figure*} 

\subsection{Using Boundaries: Fluid Settling In a Tank}\label{subsec:FluidInTank}
One method to control a swarm's shape in a bounded workspace is to simply push in a given direction until the swarm conforms to the boundary.
\paragraph{Square workplace}
This section examines the mean $(\bar{x},\bar{y})$, covariance $(\sigma^2_x,\sigma^2_y,\sigma_{xy})$, and correlation $\rho_{xy}$ of a very large swarm of robots as they move inside a square workplace under the influence of gravity pointing in the direction $\beta$. The swarm is large, but the robots are small in comparison, and together occupy a constant area $A$. Under a global input such as gravity, they flow like water, moving to a side of the workplace and forming a polygonal shape, as shown in Fig.~\ref{fig:SquareFill}. 

The range for the global input angle $\beta $ is [0,2$\pi $). In this range, the swarm assumes eight different polygonal shapes. The shapes alternate between triangles and trapezoids when the area $A$$<$1/2, and alternate between squares with one corner removed and trapezoids when $A$$>$1/2.

Computing means, variances, covariance, and correlation requires integrating over the region $R$ containing the swarm:  

\begin{align}
\bar{x} &=\frac{\iint_R x \,dx\,dy}{A} \label{eq:meanInSquareWorkspace}
\text{, }\qquad \bar{y}=\frac{\iint_R y \,dx\,dy}{A} \\
\sigma^2_x &=\frac{\iint_R \left(x-\bar{x}\right)^2  \,dx \,dy}{A}  \label{eq:varInSquareWorkspace}
\text{, } \sigma^2_y =\frac{\iint_R  \left(y-\bar{y}\right)^2 \,dx \,dy}{A}\\
\sigma_{xy} &= \frac{\iint_R  \left(x-\bar{x}_x\right) \left(y-\bar{y}\right) \, dx \,dy}{A} \label{eq:covAndcorrInSquareWorkspace}
\text{, }\rho_{xy} = \frac{\sigma^2_x}{\sigma_x\sigma_y}
\end{align}

The region of integration $R$ is the polygon containing the swarm. If the force angle is $\beta$, the mean when the swarm is in the lower-left corner is:
\begin{align}\label{eq:meanInSquareWorkspaceLL}
\bar{x}(A,\beta) &= \frac{\int_0^{\sqrt{2} \sqrt{-A \tan (\beta )}} \left(\int_0^{\sqrt{2} \sqrt{-A \cot (\beta )}+x \cot (\beta )} x \, dy\right) \, dx}{A} \nonumber \\
	&=\frac{1}{3} \sqrt{2} \sqrt{A \tan (\beta )}\\
\bar{y}(A,\beta) &= \frac{\int_0^{\sqrt{2} \sqrt{-A \tan (\beta )}} \left(\int_0^{\sqrt{2} \sqrt{-A \cot (\beta )}+x \cot (\beta )} y \, dy\right) \, dx}{A} \nonumber\\
	&=\frac{1}{3} \sqrt{2} \sqrt{A \cot (\beta )}
\end{align}
The full equations are included in the appendix, and are summarized in Fig.~\ref{fig:SquareFill}. A few highlights are that the correlation is maximized when the swarm is in a triangular shape, and is $\pm$1/2. The covariance of a triangle is always $\pm(A/18)$. Variance is minimized in the direction of $\beta$ and maximized orthogonal to $\beta$ when the swarm is in a rectangular shape. The range of mean positions are maximized when $A$ is small.

\paragraph{Circular workplace}
Though rectangular boundaries are common in artificial workspaces, biological workspaces are usually rounded.
Similar calculations can be computed for a circular workspace.  The workspace is a circle centered at (0,0) with radius 1 and thus area $\pi$.
For notational simplicity, the swarm is parameterized by the global control input signal $\beta$ and the fill-level $h$.  
Under a global input, the robot swarm fills the region under a chord with area
\begin{align}
A(h) = \arccos(1-h)-(1-h) \sqrt{(2-h) h}.
\end{align}
For a circular workspace, the locus of mean positions are aligned with $\beta$ and the mean position is at radius $r(h)$ from the center:
\begin{align}
r(h) = \frac{2 (-(h-2) h)^{3/2}}{3 \left(\sqrt{-(h-2) h} (h-1)+\arccos(1-h)\right)}
\end{align}
Variance $\sigma^2_x(\beta,h)$ is maximized at $\beta = \pi/2+n \pi$ and $h\approx1.43$, while covariance is maximized at $\beta = \pi3/4+n \pi$ and $h\approx0.92.$ For small $h$ values, correlation approaches $\pm1$. Results are summarized in Fig.~\ref{fig:CircleFill}.

%

\subsection{Using Boundaries: Friction and Boundary Layers}\label{subsec:WallFriction}
Global inputs move a swarm uniformly.  
Controlling covariance requires breaking this uniform symmetry.  A swarm inside an axis-aligned rectangular workspace can reduce variance normal to a wall by simply pushing the swarm into the boundary. Directly controlling covariance by pushing the swarm into a boundary requires changes to the boundary.  An obstacle in the lower-right corner is enough to generate positive covariance.  Generating both positive and negative covariance requires additional obstacles.  Requiring special obstacle configuration also makes covariance control dependent on the local environment. 
  Instead of pushing our robots directly into a wall, this paper examines an oblique approach, by using boundaries that generate friction with the robots.  These frictional forces are  sufficient to break the symmetry caused by uniform inputs.  Robots touching a wall have a negative friction force that opposes movement along the boundary.  This  causes robots along the boundary to slow down compared to robots in free-space. 
  
Let the control input be a vector force $\vec{F}$ with magnitude $F$ and orientation $\theta$ with respect to a line perpendicular to and into the nearest boundary. $N$ is the normal or perpendicular force between the robot and the boundary. The force of friction $F_f$ is nonzero if the robot is in contact with the boundary and  $|\theta| < \pi/2$. The resulting net force on the robot, $F_{\text{\emph{forward}}}$, is aligned with the wall and given by

\begin{align}
F_{\text{\emph{forward}}} &=  F \sin(\theta) - F_f  \nonumber \\
\text{where }  F_f &= \begin{cases}  \mu_f N, &  \mu_f N < F \sin(\theta)  \label{eq:frictionmodel}  \\
F \sin(\theta), & \text{else} \end{cases} \\ 
\text{and } N &= F \cos(\theta) \nonumber
\end{align}
 Fig.~\ref{fig:friction} shows the resultant forces on two robots when one is touching a wall. As illustrated, both experiences different net forces although each receives the same inputs.
  For ease of analysis, the following algorithms assume $\mu_f$ is infinite and robots touching the wall are prevented from sliding along the wall.
This means that if one robot is touching the wall and another robot is free, if the control input is parallel or into the wall, the touching robot will not move. 
There are many alternate models of friction that also break control symmetry. Fig.~\ref{fig:friction}c shows fluid flow along a boundary.  Fluid in the free-flow region moves uniformly, but flow decreases to zero in the boundary layer.  
\begin{align}
u(y) = u_0 [1- \frac{(y-h)^2}{h^2} = u_0 \frac{y}{h} [2- \frac{y}{h}] \label{eq:boundarylayerflow}
\end{align}

The next section shows how a system with friction model \eqref{eq:frictionmodel} and two orthogonal walls can arbitrarily position two robots. 
\begin{figure}[h]
\begin{center}
\includegraphics[width=0.8\columnwidth]{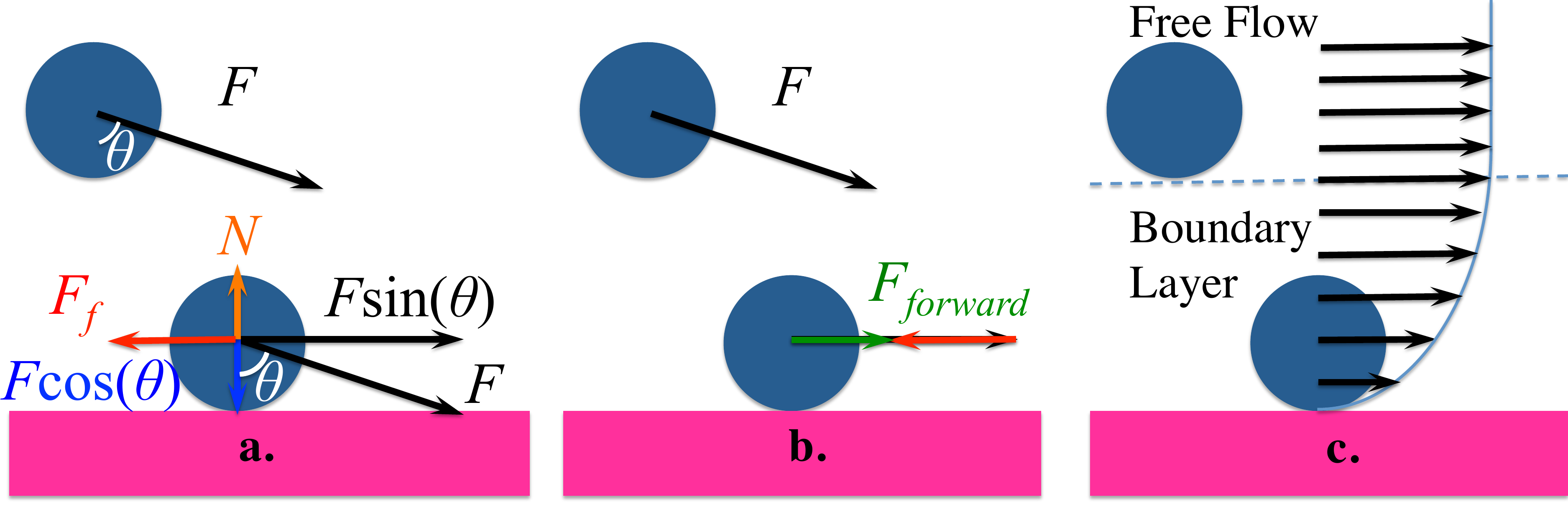} 
\vspace{-1em}
\caption{(a,b) Wall friction reduces the force for going forward $F_{\text{\emph{forward}}}$ on a robot near a wall, but not for a free robot. (c) velocity of a fluid reduces to zero at the boundary.}
\label{fig:friction}
\end{center}
\end{figure}

\section{Algorithms}\label{sec:algorithms}
\subsection{Position Control of $2$ Robots Using Wall Friction}\label{sec:PostionControl2Robots}
\begin{figure*}
\centering
\renewcommand{\figwid}{0.4\columnwidth}
{\begin{overpic}[width =\figwid]{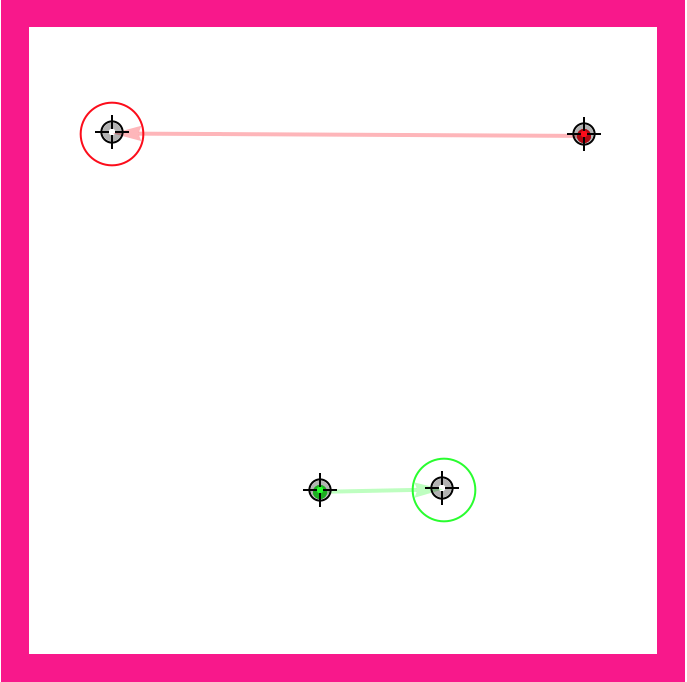}\put(10,15){$t$ = 0 s}
\end{overpic}
\begin{overpic}[width =\figwid]{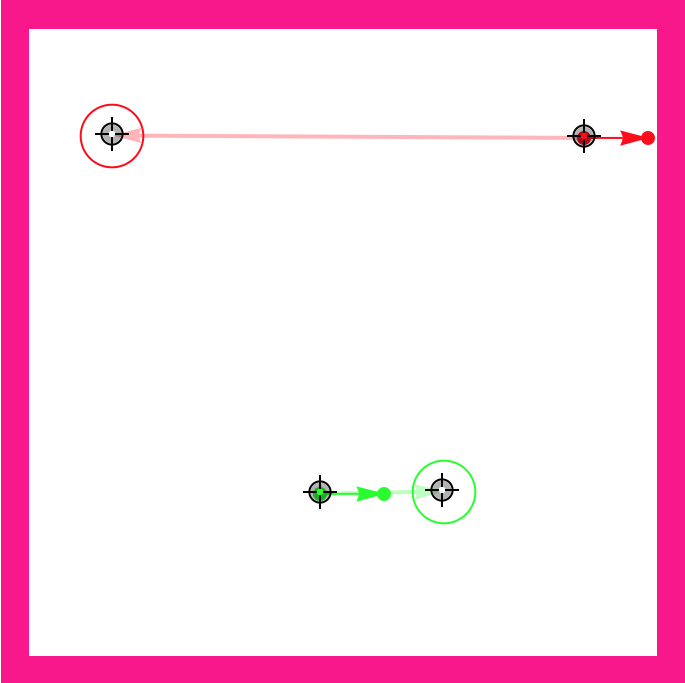}\put(10,15){$t$ = 0.14 s}
\end{overpic}
\begin{overpic}[width =\figwid]{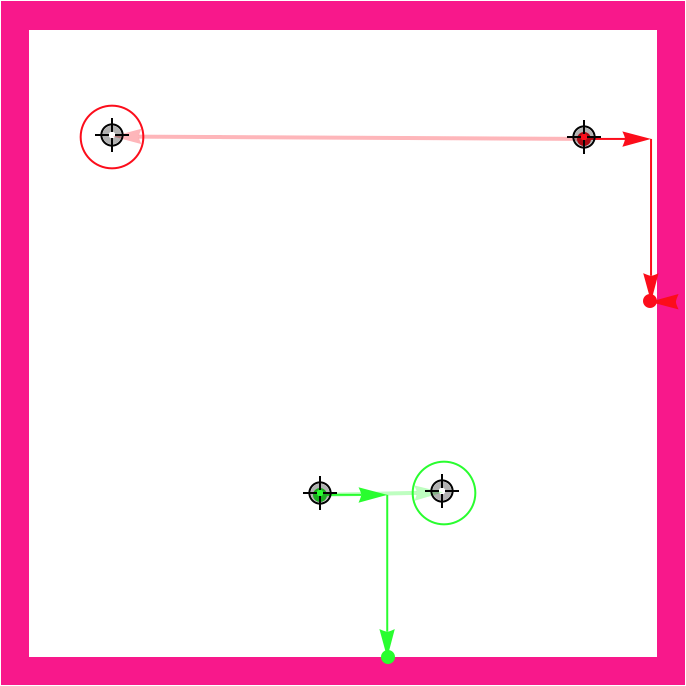}\put(10,15){$t$  = 0.29 s}
\end{overpic}
\begin{overpic}[width =\figwid]{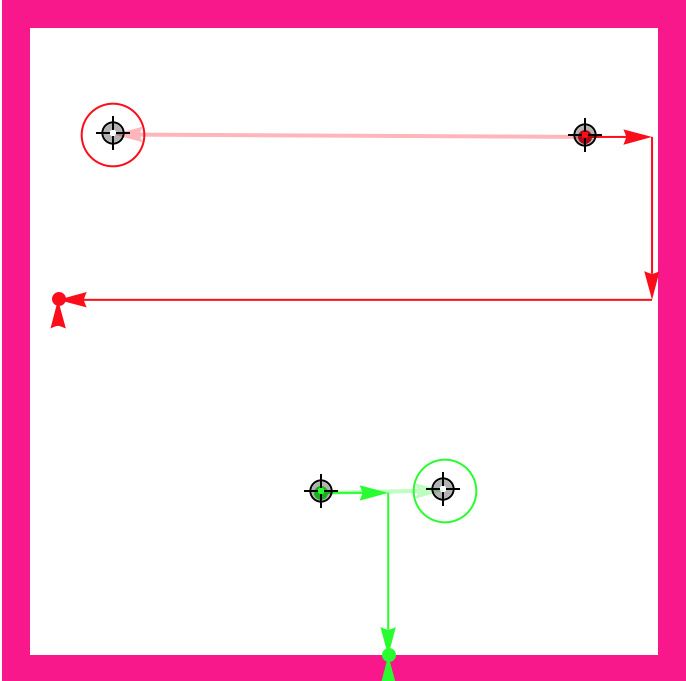}\put(10,15){$t$  = 0.43 s}
\end{overpic}
\begin{overpic}[width =\figwid]{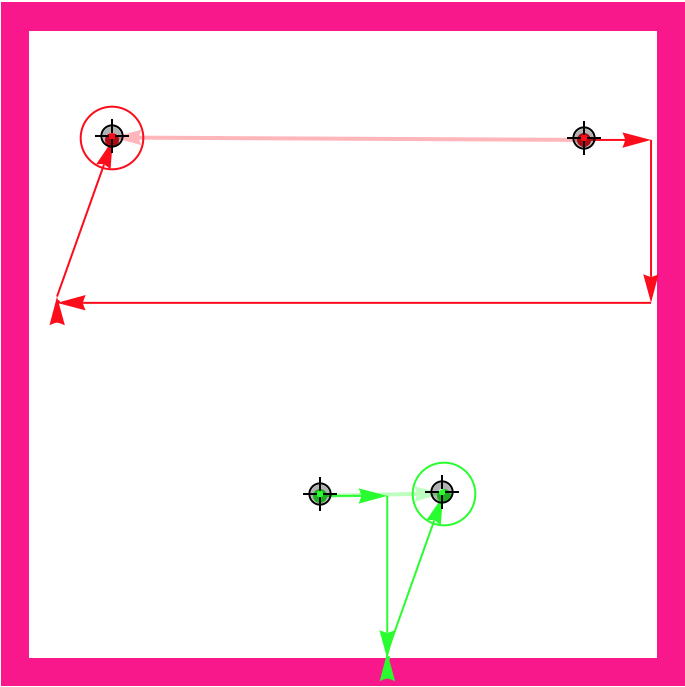}\put(10,15){$t$  = 0.76 s}
\end{overpic}}\\
\caption{\label{fig:shapeControlMathematica1}{Frames from an implementation of Alg.\ \ref{alg:PosControl2Robots}: two robot positioning using walls with infinite friction. The algorithm only requires friction along the bottom and left walls.
Robot initial positions are shown by a crosshair, and final positions by a circled crosshair.  Dashed lines show the shortest route if robots could be controlled independently.  Solid arrows show path given by  Alg.\ \ref{alg:PosControl2Robots}.
Online demonstration and source code at \citep{Shahrokhi2015mathematicaParticle}.
}
}
\end{figure*}

This section describes Alg.~\ref{alg:PosControl2Robots}, which uses wall-friction to arbitrarily position two robots in a rectangular workspace.  This algorithm  introduces concepts that will be used for multi-robot positioning. It only requires collisions with two orthogonal walls, in this case, the bottom and left walls. Fig.~\ref{fig:shapeControlMathematica1} shows a Mathematica implementation of the algorithm, and is useful as a visual reference for the following description.

Assume two robots are initialized at $s_1$ and $s_2$ with corresponding goal destinations $e_1$ and $e_2$. 
Denote the current positions of the robots  $r_1$ and $r_2$. 
Subscripts $_x$ and $_y$ denote the $x$ and $y$ coordinates, i.e., $s_{1x}$ and $s_{1y}$ denote the $x$ and $y$ locations of $s_1$. 
The algorithm assigns a global control input at every instance.
The goal is to adjust 
 $\Delta r_x = r_{2x}-r_{1x}$ from $\Delta s_x = s_{2x}-s_{1x}$ to $\Delta e_x = e_{2x}-e_{1x}$ and  adjust 
 $\Delta r_y = r_{2y}-r_{1y}$ from $\Delta s_y = s_{2y}-s_{1y}$ to $\Delta e_y = e_{2y}-e_{1y}$ using a shared global control input. 
 This algorithm exploits the position-dependent friction model \eqref{eq:frictionmodel}.

Our algorithm solves the positioning problem in two steps: 
First, $|\Delta r_x - \Delta e_x |$ is reduced to zero while  $\Delta r_y$ is kept constant in Alg.~\ref{alg:XControl}. 
Second $|\Delta r_y - \Delta e_y |$ is reduced to zero while  $\Delta r_x$ is kept constant.

\begin{algorithm}
\caption{WallFrictionArrange2Robots($s_1,s_2,e_1,e_2,L$)}\label{alg:PosControl2Robots}
\begin{algorithmic}[1]
\Require starting $(s_1,s_2)$ and ending $(e_1,e_2)$ positions of  two robots. 
$(0,0)$ is bottom corner, $s_1$ is rightmost robot, 
 $L$ is length of the walls. 
 Current position of the robots are $(r_1,r_2)$.

\State ($r_1,r_2$) = GenerateDesired$x$-spacing($s_1,s_2,e_1,e_2,L$)
\State GenerateDesired$y$-spacing($r_1,r_2,e_1,e_2,L$)

\end{algorithmic}
\end{algorithm}

\begin{algorithm}
\caption{GenerateDesired$x$-spacing($s_1,s_2,e_1,e_2,L$)}\label{alg:XControl}
\begin{algorithmic}[1]
\Require Knowledge of starting $(s_1,s_2)$ and ending $(e_1,e_2)$ positions of  two robots. 
$(0,0)$ is bottom corner, $s_1$ is topmost robot, 
 $L$ is length of the walls. Current robot positions are $(r_1,r_2)$.
\Ensure   $ r_{1y} - r_{2y}  \equiv s_{1y} - s_{2y} $   
\State $\epsilon \gets $ small number
\State $ \Delta s_x  \gets s_{1x} - s_{2x} $
\State $ \Delta e_x \gets e_{1x} - e_{2x} $
\State $ r_1 \gets s_1$, $ r_2 \gets s_2$
\If {$\Delta e_x < 0 $ }
\State $ m \gets ( L-\epsilon-\max( r_{1x},r_{2x}) ,0)   $ \Comment{Move to right wall}
\Else 
\State  $ m \gets ( \epsilon-\min( r_{1x},r_{2x}),0 )    $ \Comment{Move to left wall}
\EndIf
\State $m  \gets  m + (0, -\min( r_{1y},r_{2y} ))$ \Comment{Move to bottom}
\State $ r_1 \gets r_1+m$, $ r_2 \gets r_2+m$ \Comment{Apply move}
\If {$\Delta e_x - (r_{1x} - r_{2x} ) > 0 $}
\State $ m \gets (\min(|\Delta e_x - \Delta s_x |, L- r_{1x}), 0)$  \Comment{Move right}
\Else
\State $ m \gets (-\min(|\Delta e_x - \Delta s_x |, r_{1x}), 0)$\Comment{Move left}
\EndIf 
\State $m  \gets  m + (0, \epsilon)$ \Comment{Move up}
\State $ r_1 \gets r_1+m$, $ r_2 \gets r_2+m$ \Comment{Apply move}
\State $\Delta r_x = r_{1x} - r_{2x}$
\If {$\Delta r_x \equiv \Delta e_x$} 
\State  \Return $(r_1,r_2)$
\Else   
\State \Return GenerateDesired$x$-spacing($r_1,r_2,e_1,e_2,L$)
\EndIf
\end{algorithmic}
\end{algorithm}


\subsection{Position Control of $n$ Robots Using Wall Friction}\label{sec:PostionControlnRobots}
\begin{figure}
\begin{center}
	\includegraphics[width=1.0\columnwidth]{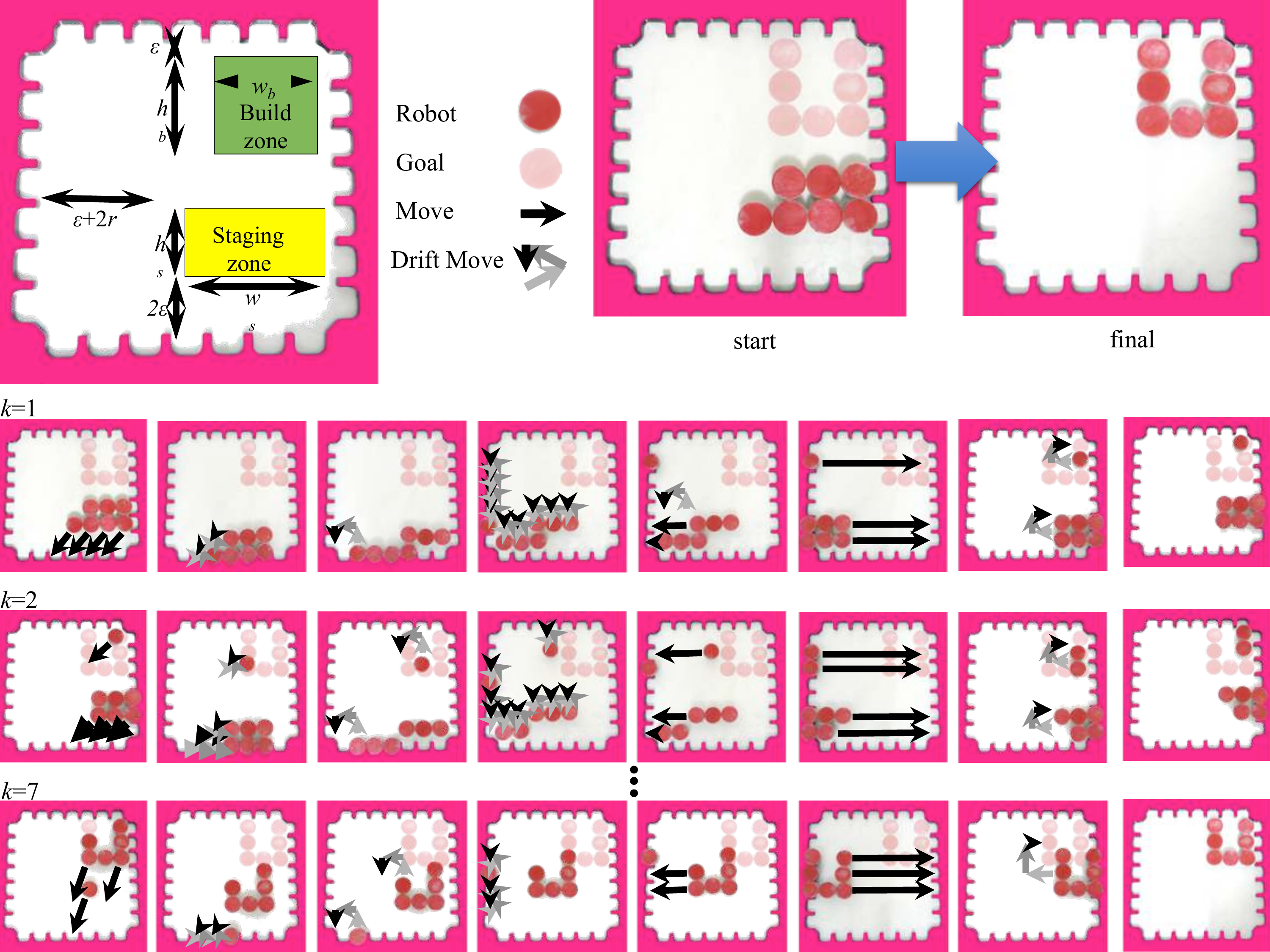}
\end{center}
\vspace{-1em}
\caption{\label{fig:construction2d}
Illustration of Alg.\ \ref{alg:PosControlNRobots}, $n$ robot position control  using wall friction.
}
\end{figure}

Alg. \ref{alg:PosControl2Robots}  can be extended to control the position of $n$ robots using wall friction under several constraints. The solution described here is an iterative procedure with $n$ loops. The $k$th loop moves the $k$th robot from a \emph{staging zone} to the desired position in a \emph{build zone}. All robots move according to the global input, but due to wall friction, at the end the $k$th loop, robots 1 through $k$ are in their desired final configuration in the build zone, and robots $k+1$ to $n$ are in the staging zone. See Fig.~\ref{fig:construction2d} for a schematic of the build and staging zones.

\begin{figure}
\begin{center}
	\includegraphics[width=.9\columnwidth]{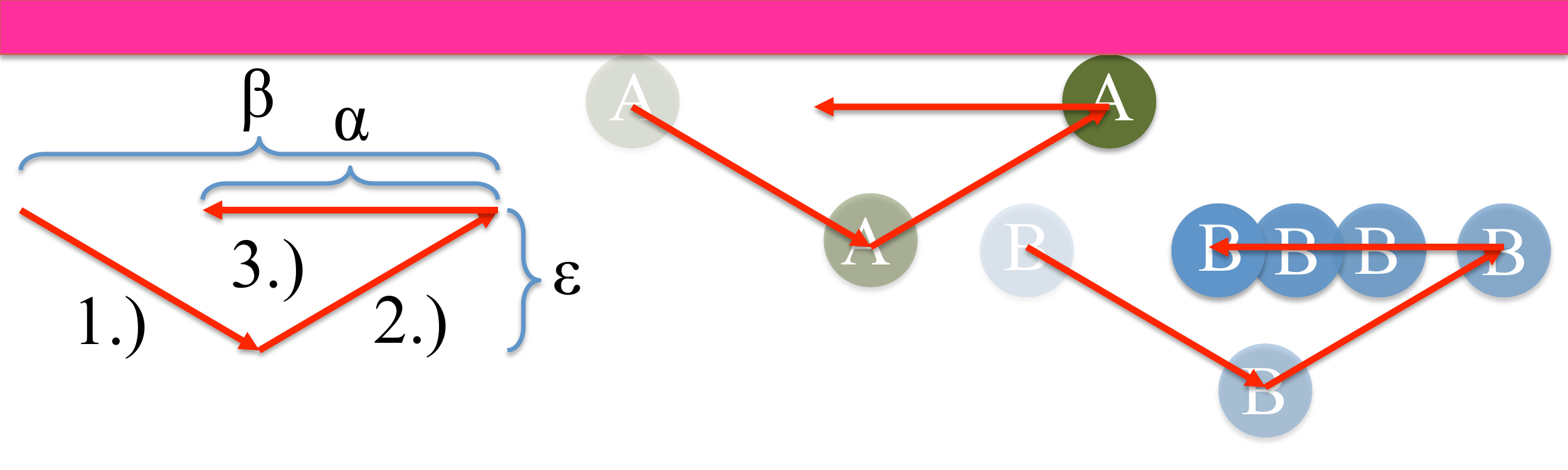}
\end{center}
\vspace{-1em}
\caption{\label{fig:driftmove}
A  $\operatorname{DriftMove}(\alpha, \beta, \epsilon)$ to the right repeats a triangular movement sequence $\{ (\beta/2,-\epsilon),(\beta/2,\epsilon),(-\alpha,0)\}$. Robot $A$ touching a top wall moves right $\beta$ units, while robots not touching the top move right $\beta-\alpha$.
}
\end{figure}

Assume an open workspace with four axis-aligned walls with infinite friction.
The axis-aligned build zone of dimension $(w_b, h_b)$ containing the final configuration of $n$ robots must be disjoint from the axis-aligned staging zone of dimension $(w_s, h_s)$  containing the starting configuration of $n$ robots. Without loss of generality, assume the build zone  is above the staging zone. 
Furthermore, there must be at least $\epsilon$ space above the build zone, $\epsilon$ below the staging zone, and $\epsilon + 2r$ to the left of the build and staging zone, where $r$ is the radius of a robot.  The minimum workspace is then $(\epsilon + 2r + \max(w_f,w_s), 2\epsilon + h_s,h_f)$.

The $n$ robot position control algorithm relies on a $\operatorname{DriftMove}(\alpha, \beta, \epsilon)$ control input, shown in Fig.\  \ref{fig:driftmove}.
A drift move consists of repeating a triangular movement sequence $\{ (\beta/2,-\epsilon),(\beta/2,\epsilon),(-\alpha,0)\}$. The robot touching a top wall moves right $\beta$ units, while robots not touching the top move right $\beta-\alpha$.

Let $(0,0)$ be the lower left corner of the workspace, $p_k$ the $x,y$ position of the $k$th robot, and $f_k$ the final $x,y$ position of the $k$th robot. Label the robots in the staging zone from left-to-right and top-to-bottom, and the $f_k$ configurations right-to-left and top-to-bottom as shown in Fig.~\ref{fig:construction2d}.

\begin{algorithm}
\caption{PositionControl$n$RobotsUsingWallFriction($k$)}\label{alg:PosControlNRobots}
\begin{algorithmic}[1]
\State move( $-\epsilon, r-p_{k,y}$) 

\While{ $p_{k,x} > r$} 
\State $\operatorname{DriftMove}(\epsilon, \min(p_{k,x} - r,\epsilon), \epsilon)$ left   
\EndWhile

\State $m \gets \operatorname{ceil}(\frac{f_{k,y}-r}{\epsilon})$
\State $\beta \gets \frac{f_{k,y}-r}{m}$
\State $\alpha \gets \beta - \frac{r - p_{k,y}-\epsilon}{m}$
\For{ $m$ iterations}
\State $\operatorname{DriftMove}(\alpha, \beta, \epsilon)$ up   
\EndFor

\State move($r+\epsilon-f_{k,x}, 0$)  
\State move($f_{k,x}-r, 0$)  

\end{algorithmic}
\end{algorithm}

Alg. \ref{alg:PosControlNRobots} proceeds as follows:  
First, the robots are moved left away from the right wall, and down so robot $k$ touches the bottom wall.
Second, a set of $\operatorname{DriftMove()}$s are executed that move robot $k$ to the left wall with no net movement of the other robots.
Third, a set of $\operatorname{DriftMove()}$s are executed that  move robot $k$ to its target height and return the other robots to their initial heights. 
Fourth, all robots except robot $k$ are pushed left until robot $k$ is in the correct relative $x$ position compared to robots 1 to $k-1$.
Finally, all robots are moved right until robot $k$ is in the desired target position.

\subsection{Controlling Covariance Using Wall Friction}\label{subsec:ClosedLoopCovarianceControl}
Assume an open workspace with infinite boundary friction. Goal variances and covariance are  $(\sigma_{goalx}^2,\sigma_{goaly}^2, \sigma_{goalxy}$ and mean, variances and covariance of the swarm  are $( \bar{x},\bar{y},\sigma_x^2,\sigma_y^2, \sigma_{xy})$. For our experiments, $c_1 = 0.1$.
\begin{enumerate}
\item swarm is pushed into the left wall until $\sigma_x^2< c_1\sigma_{goalx}^2$.  
\item swarm's mean position is moved to the center of the workspace
\item swarm is pushed into the bottom wall until $\sigma_y^2 \le \sigma_{goalt}^2$. 
\item if $\sigma_{goalxy}>0$ swarm slides right until $\sigma_{xy} \ge \sigma_{goalxy}$ \\
else swarm slides left until $\sigma_{xy} \le \sigma_{goalxy}$ 
\item  swarm's mean position is moved to the center of the workspace
\end{enumerate}

%


\section{Simulation}\label{sec:simulation}

Two simulations were implemented using wall-friction for position control.  The first controls the position of two robots, the second controls the position of $n$ robots.  All code is available online at \citet{Zhao2016mathematica,Zhao2016mathematicaSquare}.

Two additional simulations were performed using wall-friction to control variance and covariance.  The first is an open-loop algorithm that demonstrates the effect of varying friction levels.  The second uses a closed-loop controller to achieve desired variance and covariance values.

\subsection{Position Control of Two Robots}

Algorithms \ref{alg:PosControl2Robots}, \ref{alg:XControl}, 
were implemented in Mathematica using point robots (radius = $0$).  Fig.~\ref{fig:shapeControlMathematica1}  shows  this algorithm for two configurations. 
Robot initial positions are shown by a crosshair, and final positions by a circled crosshair.  Dashed lines show the shortest route if robots could be controlled independently.  The path given by  Alg.\ \ref{alg:PosControl2Robots} is shown with solid arrows.
Each row has five snapshots taken every quarter second. For the sake of brevity axis-aligned moves were replaced with oblique moves that combine two moves simultaneously. 
 $\Delta r_x$ is adjusted to $\Delta e_x$ in the second snapshot at $t = 0.25$. 
 The following frames  adjust $\Delta r_y$ to $\Delta e_y$. 
 $\Delta r_y$ is corrected by $t = 0.75$. 
 Finally, the algorithm moves the robots to their corresponding destinations.

In the worse case, adjusting both $\Delta r_x$ and $\Delta r_y$ requires two iterations.   Two iterations of Alg. \ref{alg:XControl} are only required if $|\Delta e_x - \Delta s_x|>L$. 
Similarly,  two iterations 
are only required if $|\Delta e_y - \Delta s_y|>L$. An online interactive demonstration and source code of the algorithm are available at \citep{Shahrokhi2015mathematicaParticle}.


\subsection{Position Control of $n$ Robots}
\begin{figure}
\begin{center}
	\includegraphics[width=\columnwidth]{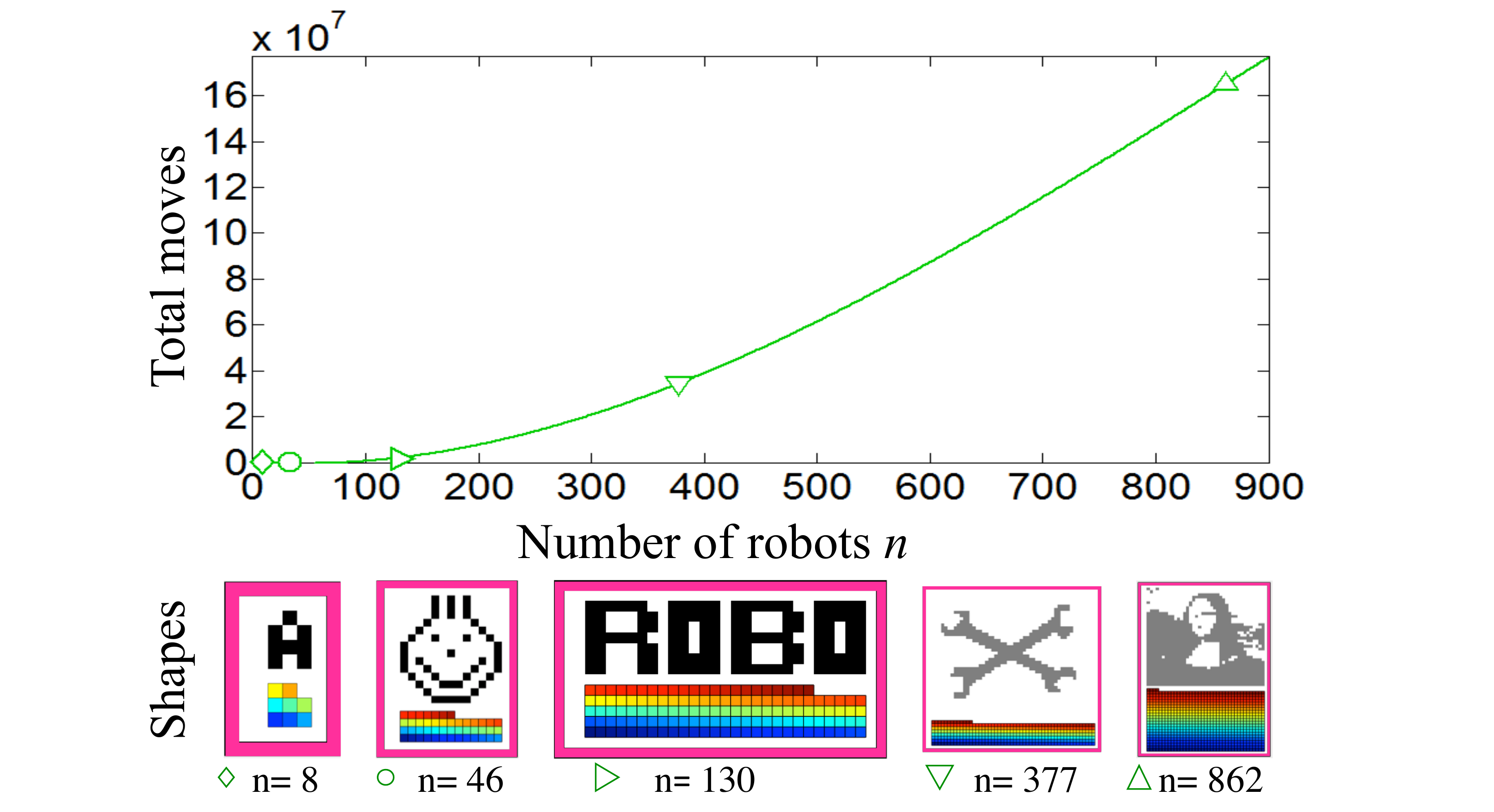}
\end{center}
\vspace{-1em}
\caption{\label{fig:4diagramsplots.pdf}
The required number of moves under Alg. \ref{alg:PosControlNRobots}  using wall-friction to rearrange $n$ square-shaped 
robots.  
See hardware implementation and simulation at \citep{Arun2015}.
}
\end{figure}

Alg. \ref{alg:PosControlNRobots}  was simulated in {\sc Matlab} using square block robots with unity width. Code is available at \citep{Arun2015}.
Simulation results are shown in Fig.~\ref{fig:4diagramsplots.pdf} for arrangements with an increasing number of robots,  $n$= [8, 46, 130, 390, 862]. 
The distance moved grows quadratically with the number of robots $n$. A best-fit line $210 n^2 + 1200n-10,000$ is overlaid by the data..

\begin{figure}
\begin{center}
	\includegraphics[width=\columnwidth]{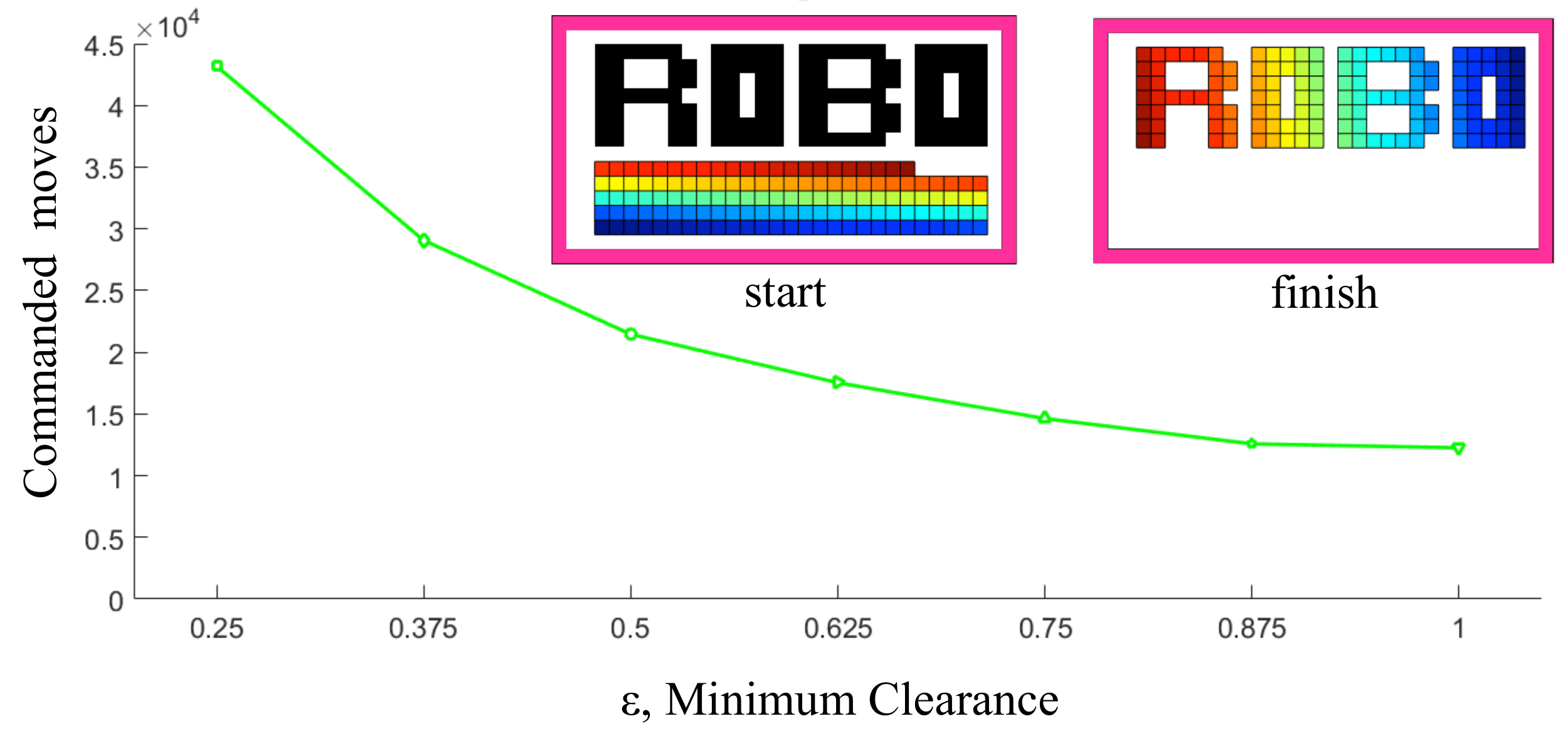}
\end{center}
\vspace{-1em}
\caption{\label{fig:graphrobo.pdf}
Control performance is sensitive to the desired clearance $\epsilon$.  As $\epsilon$ increases, the total distance decreases asymptotically.
}
\end{figure}

In Fig.~\ref{fig:4diagramsplots.pdf}, the amount of clearance is $\epsilon=1$.
Control performance is sensitive to the desired clearance.  As $\epsilon$ increases, the total distance decreases asymptotically, as shown in Fig.~\ref{fig:graphrobo.pdf}, because the robots have more room to maneuver and fewer $\operatorname{DriftMove}$s are required.


\subsection{Efficient Control of Covariance}
\begin{figure}
\begin{center}
	\includegraphics[width=\columnwidth]{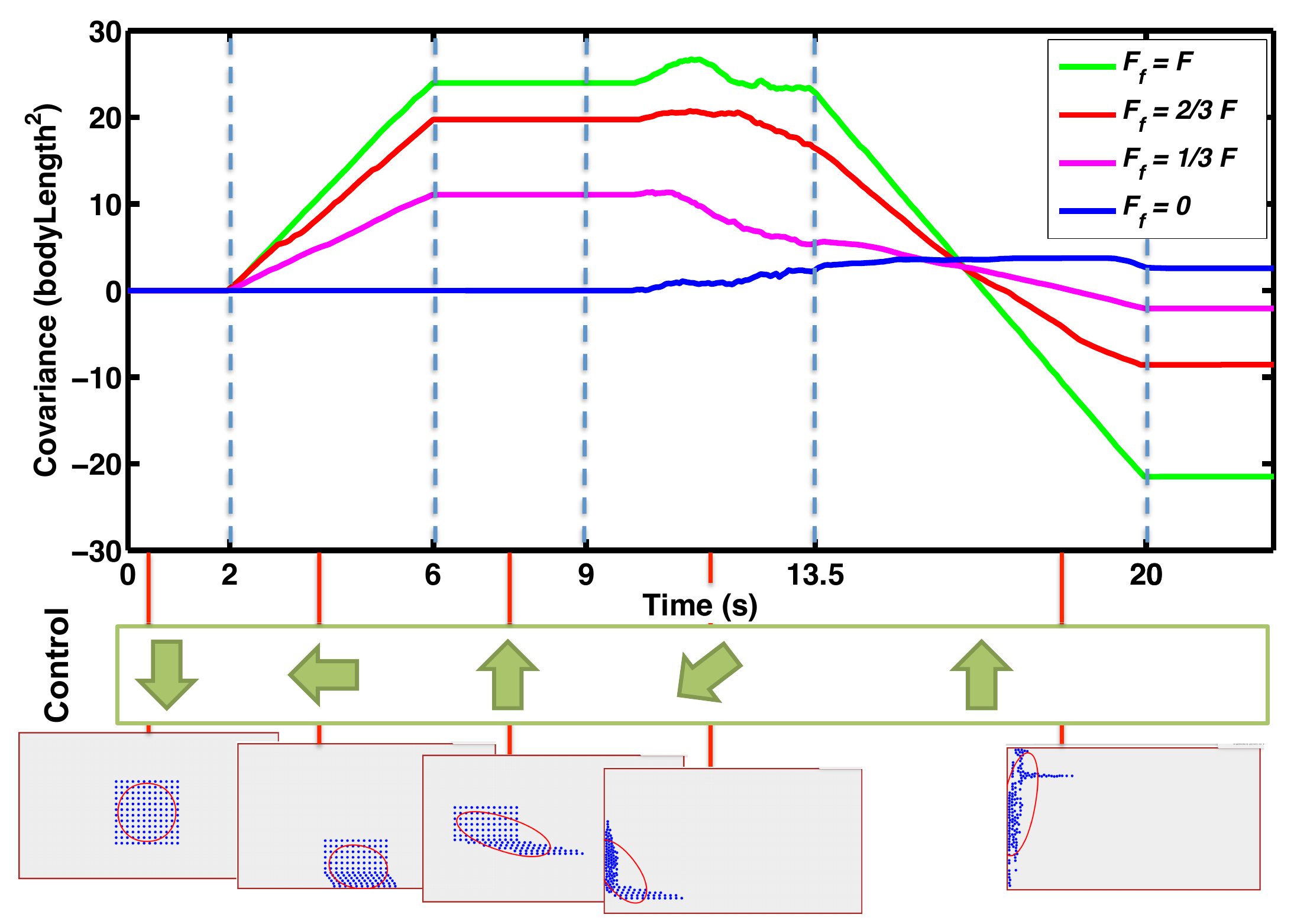}
\end{center}
\vspace{-1em}
\caption{\label{fig:SimCovarianceFuncFrictionOpenLoop}
Open-loop simulation with 144 disc robots and varying levels of boundary friction under the same initial conditions.  Without friction, covariance is unchangeable.  As friction increases, the covariance can be manipulated to greater degrees.
}
\end{figure}

Random disturbances impair the performance of Alg. \ref{alg:PosControl2Robots} and Alg. \ref{alg:PosControlNRobots}. Still, we are able to control covariance of the swarm. This section demonstrates simulations of controlling covariance of the swarm.
These simulations use  the 2D physics engine Box2D, by \citet{catto2010box2d}.
 144 disc-shaped robots were controlled by an open-loop control input as illustrated in Fig. \ref{fig:SimCovarianceFuncFrictionOpenLoop}.  All robots had  the same initial conditions, but in four tests the boundary friction was $F_f = \{0,1/3 F, 2/3F, F\}$.
 Without friction, covariance has minimal variation.  As friction increases, the covariance can be manipulated to greater degrees.

\begin{figure}
\begin{center}
	\includegraphics[width=1.0\columnwidth]{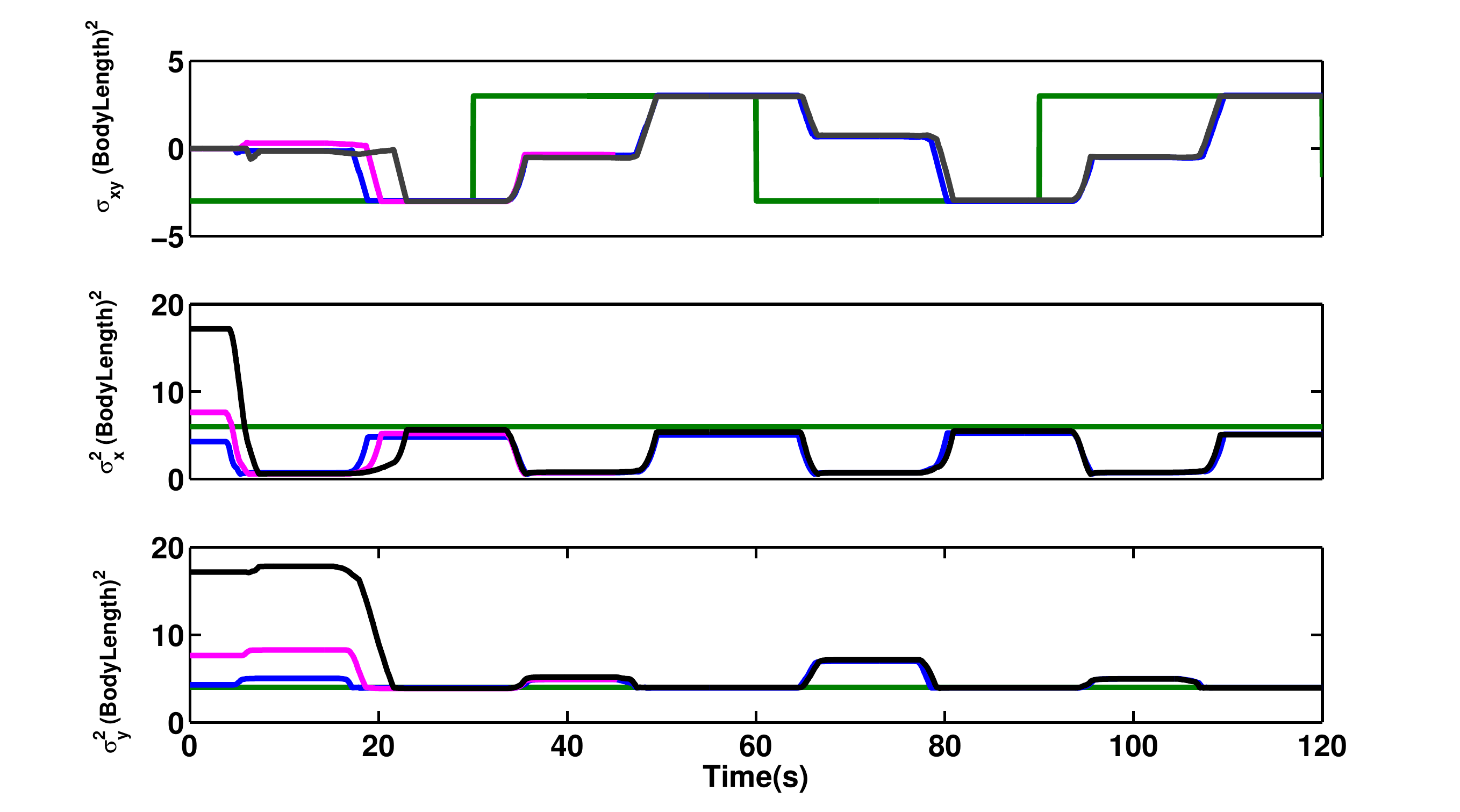}
\end{center}
\vspace{-1em}
\caption{\label{fig:CovVarControlPlot}
Closed-loop simulation with 144 disc robots and three sets of initial conditions.  The algorithm tracks goal variance and covariance values (green).  The goal covariance switches sign every 30 s.}
\end{figure}

 144 disc-shaped robots were also controlled by a closed-loop controller using the procedure in \S \ref{subsec:ClosedLoopCovarianceControl}. Fig. \ref{fig:CovVarControlPlot} illustrates that covariance and variances in $x$ and $y$ axis were controlled from a set of initial conditions.


\section{experiment}\label{sec:expResults}

\begin{figure*}[!htb]
\centering
\renewcommand{\figwid}{0.38\columnwidth}
{
\begin{overpic}[width =\figwid]{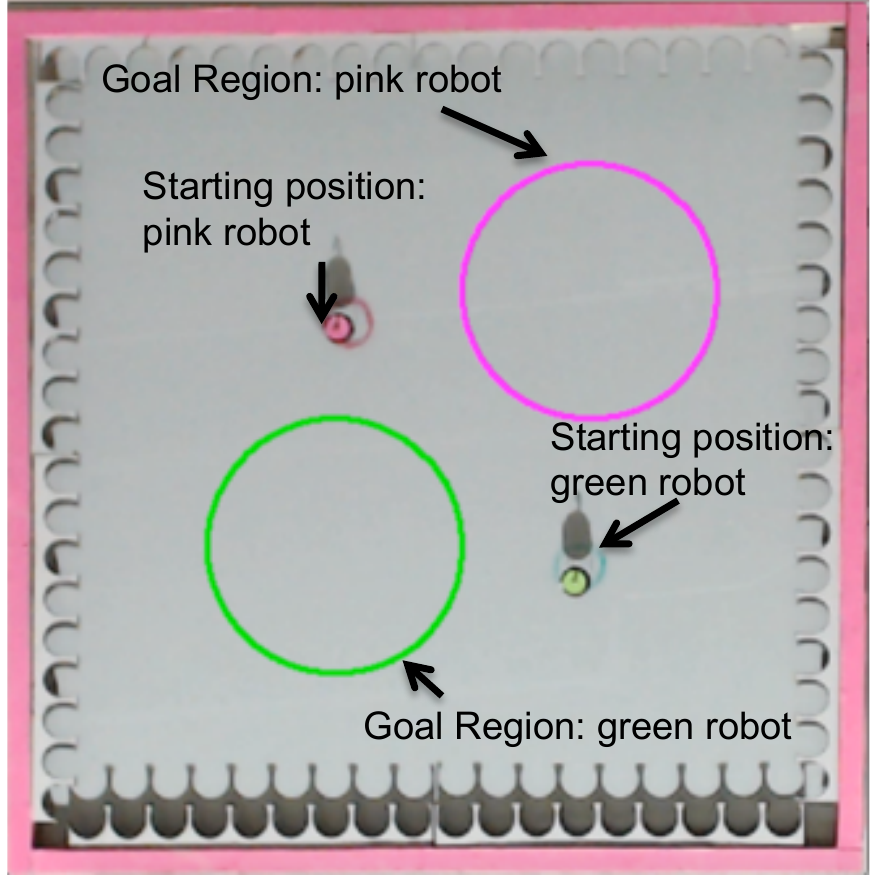}
\end{overpic}
\begin{overpic}[width =\figwid]{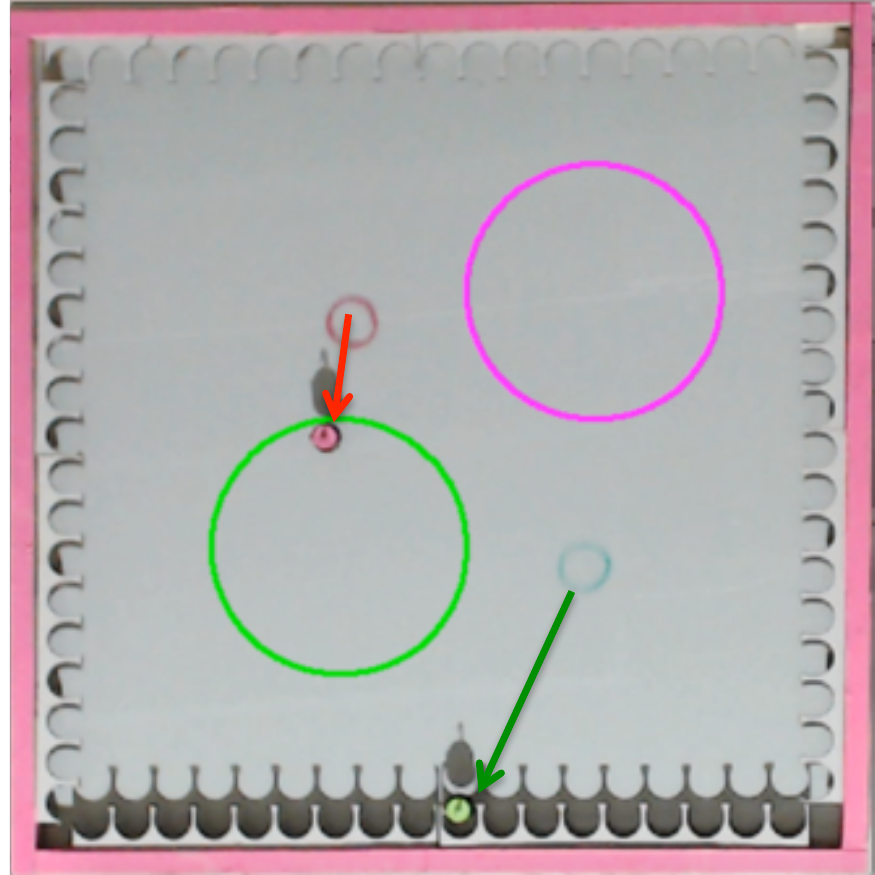}\put(12,70){$t$  = 60 s}
\end{overpic}
\begin{overpic}[width =\figwid]{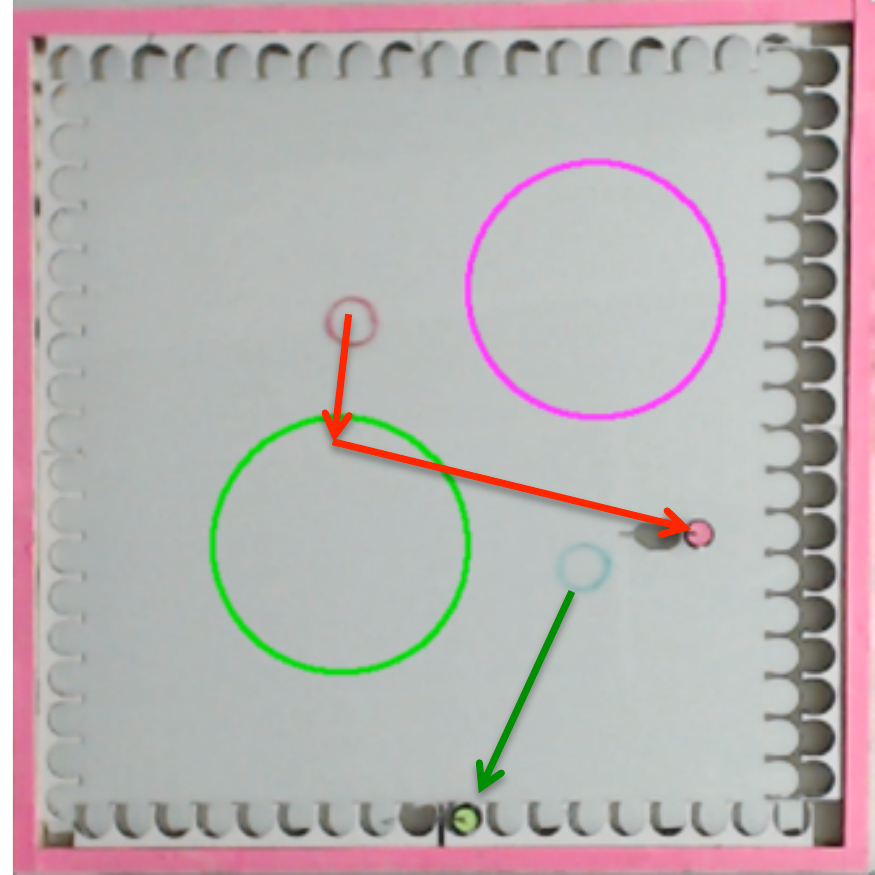}\put(12,70){$t$  = 150 s}
\end{overpic}
\begin{overpic}[width =\figwid]{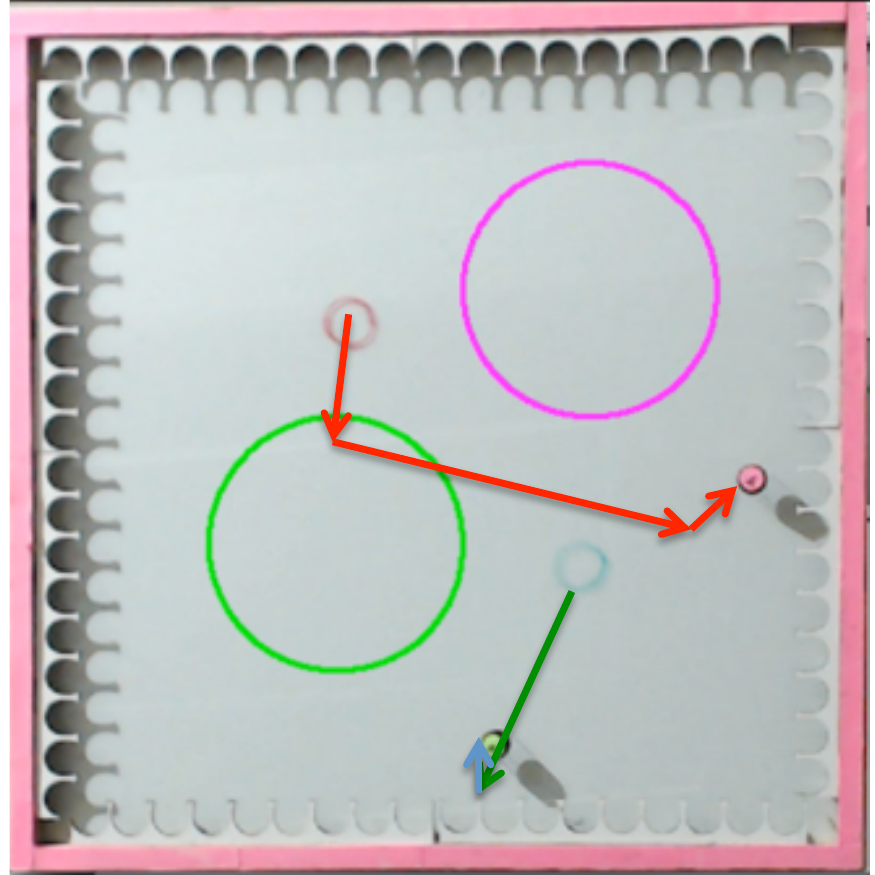}\put(12,70){$t$  = 160 s}
\end{overpic}
\begin{overpic}[width =\figwid]{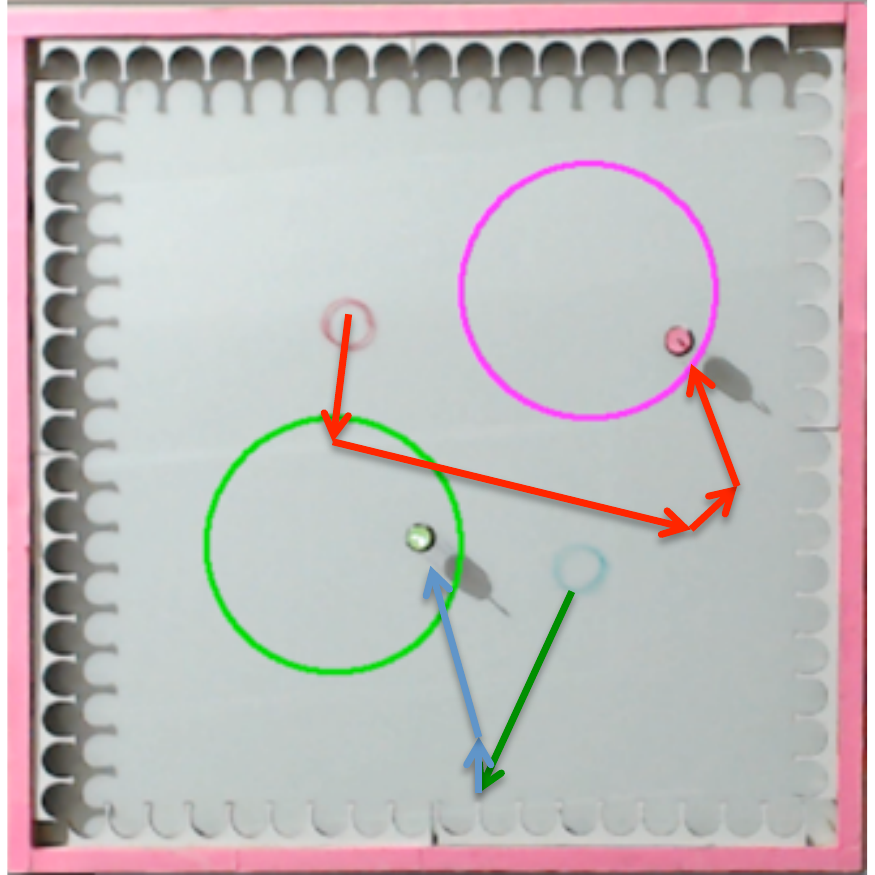}\put(12,70){$t$  = 210 s}
\end{overpic}}
\caption{\label{fig:storyReal}Position control of two kilobots (Alg. \ref{alg:XControl}) steered to corresponding colored circle. Boundary walls have nearly infinite friction, so the green robot is stopped by the wall from $t = 60$s until the commanded input is directed away form the wall at $t=150$s, while the pink robot in free-space is unhindered.}
\end{figure*}

\begin{figure}
\begin{center}
	\includegraphics[width=.9\columnwidth]{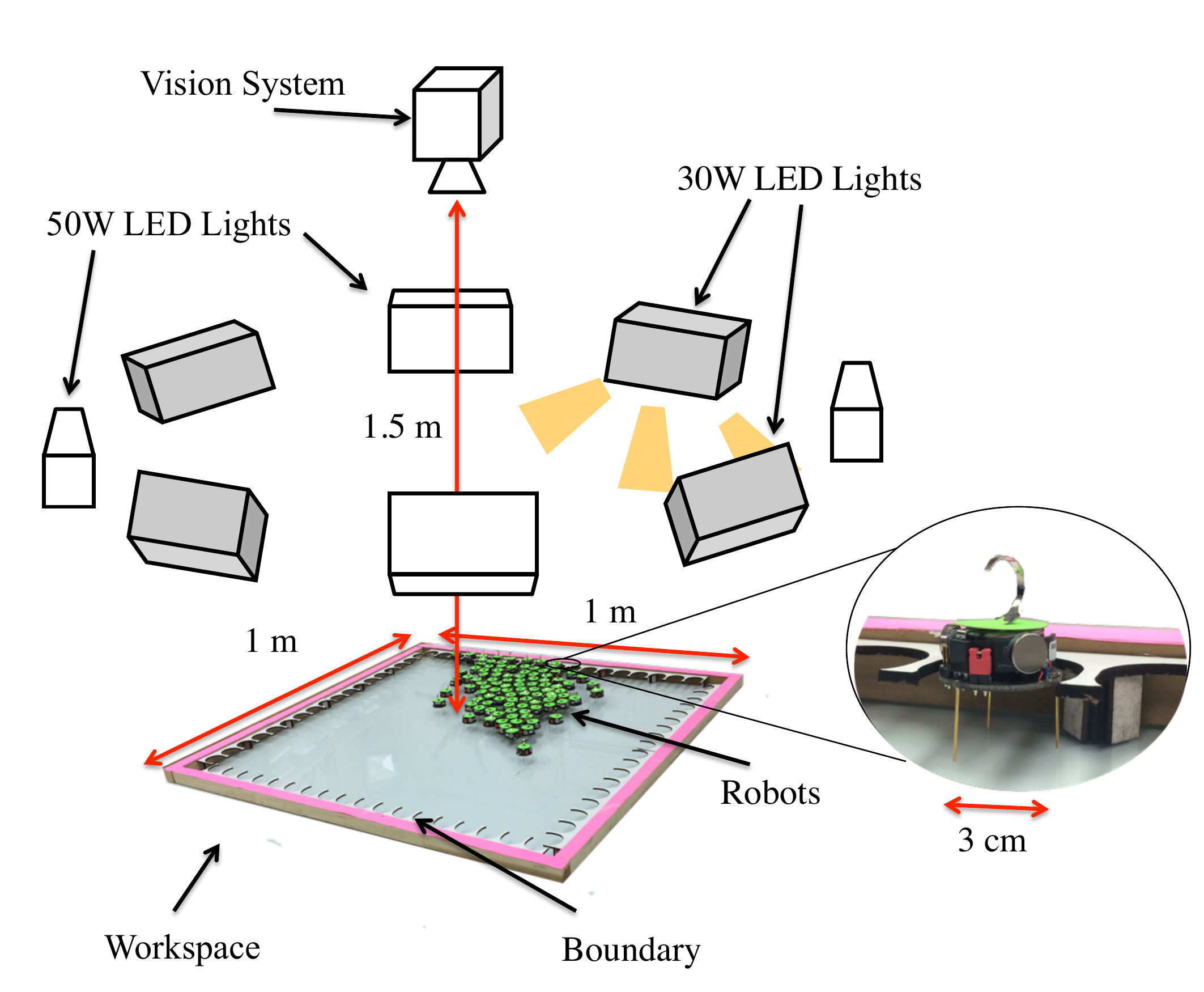}
\end{center}
\vspace{-1em}
\caption{\label{fig:setup}
Hardware platform:  table with 1$\times$1 m workspace, surrounded by eight  triggered LED floodlights, with an overhead machine vision system.
}
\vspace{-1em}
\end{figure}

Our experiments are on centimeter-scale hardware systems called \emph{kilobots}.  These allows us to emulate a variety of dynamics, while enabling a high degree of control over robot function, the environment, and data collection. The kilobot, from \citet{Rubenstein2012,rubenstein2014programmable} is a low-cost robot designed for testing collective algorithms with large numbers of robots. It is available as an open source platform or commercially from~\citet{K-Team2015}.  Each robot is approximately 3 cm in diameter, 3 cm tall, and uses two vibration motors to move on a flat surface at speeds up to 1 cm/s.  Each robot has one ambient light sensor that is used to implement \emph{phototaxis},  moving towards a light source. 
In these experiments as shown in Fig.~\ref{fig:setup}, we used $n$=100 kilobots, a 1 m$\times$1 m whiteboard as the workspace, four 30W and four 50W LED floodlights arranged 1.5 m above the plane of the table at the $\{N,NE,E,SE,S,SW,W,NW\}$ vertices of a 6 m square centered on the workspace. The lights were controlled using an Arduino Uno board connected to an 8-relay shield.  Above  the table, an overhead machine vision system tracks the position of the swarm.

\subsection{Hardware Experiment: Position Control of Two Robots}
The walls of the hardware platform have almost infinite friction, due to a laser-cut, zigzag border. When a kilobot is steered into the zigzag border, they pin themselves to the wall unless the global input directs them away from the wall.  This wall friction is sufficient to enable independent control of two kilobots, as shown in Fig.~\ref{fig:storyReal}.

\subsection{Hardware Experiment: Position Control of n Robots}
The hardware setup has a bounded platform, magnetic sliders, and a magnetic guide board.  Designs for each are available at \citep{arunhardware}. The pink boundary is toothed with a white free space, as shown in Fig \ref{fig:construction2d}. Only discrete, 1 cm moves in the $x$ and $y$ directions are used. The goal configuration highlighted in the top right corner represents a `U' made of seven sliders. The dark red configuration is the current position of the sliders. 
Due to the discretized movements allowed by the boundary, drift moves follow a 1 cm square.  Free robots return to their start positions but robots on the boundary to move laterally, generating a net sliding motion.

Fig. \ref{fig:construction2d} follows the motion of the sliders through iterations  $k$=1, 2 and 7. All robots receive the same control inputs, but boundary interactions breaks the control symmetry.  Robots reach their respective goal position in a first-in, first-out arrangement beginning with the bottom-left robot from the staging zone occupying the top-right position of the build zone.

\subsection{Hardware Experiment: Control of Covariance}
\begin{figure}
\begin{center}
	\includegraphics[width=\columnwidth]{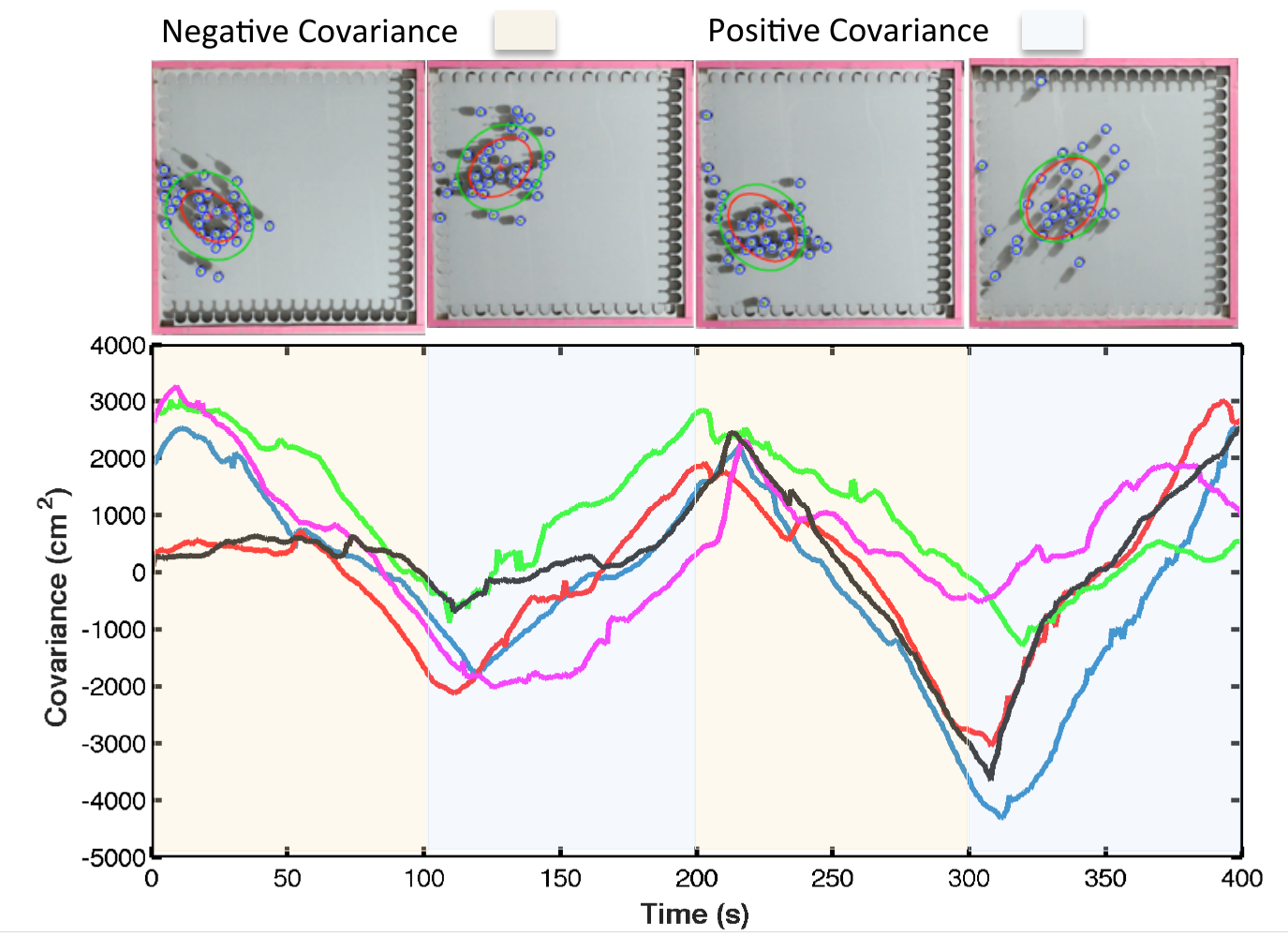}
\end{center}
\vspace{-1em}
\caption{\label{fig:covExperiment}
Hardware demonstration steering 100 kilobot robots to desired covariance. The goal covariance is negative in first 100 seconds and is positive in the next 100 seconds. The actual covariance is shown in different trials. Frames above the plot show output from machine vision system and an overlaid covariance ellipse.
\vspace{-1em}
}
\end{figure}

To demonstrate covariance control $n$=100 robots were placed on the workspace and manually steered with lights, using friction with the boundary walls to vary the covariance from  -4000 to 3000 cm$^2$.  The resulting covariance is plotted in Fig.~\ref{fig:covExperiment}, along with snapshots of the swarm.

\section{Related Work}\label{sec:RelatedWork}

Controlling the \emph{shape}, or relative positions, of a swarm of robots is a key ability for a range of applications.  Correspondingly, it has been studied from a control-theoretic perspective in  both centralized and decentralized approaches. For examples of each, see the centralized virtual leaders in \citet{egerstedt2001formation}, and the  gradient-based decentralized controllers  using control-Lyapunov functions in~\citet{hsieh2008decentralized}. However, these approaches assume a level of intelligence and autonomy in individual robots that exceeds the capabilities of many systems, including current micro- and nano-robots.  Current micro- and nano-robots, such as those in~\citet{martel2015magnetotactic,Xiaohui2015magnetiteMicroswimmers} and \citet{Chowdhury2015}, lack onboard computation.

Instead, this paper focuses on centralized techniques that apply the same control input to each member of the swarm. 
Precision control requires breaking the symmetry caused by the global input.  
Symmetry can be broken using agents that respond differently to the global control, either through agent-agent reactions, see work modeling biological swarms \citet{bertozzi2015ring}, or engineered inhomogeneity  \citet{bretl2007,Donald2013,beckerIJRR2014}.
This work assumes a uniform control \eqref{eq:swarmDynamics}  with homogenous agents, as in~\citet{Becker2013b}. 
The techniques in this paper are inspired by fluid-flow techniques and artificial force-fields. 

\emph{Fluid-flow:} 
Shear forces are unaligned forces that push one part of a body in one direction, and another part of the body in the opposite direction. 
These are common in fluid flow along boundaries.  
Most introductory fluid dynamics textbooks provide models, for example, see~\citet{Munson2013}.
Similarly, a swarm of robots under global control pushed along a boundary will experience shear forces.  
This is a position-dependent force, and so can be exploited to control the configuration or shape of the swarm.  
Physics-based swarm simulations used these forces to disperse a swarm's spatial position for coverage in \citet{spears2006physics}.

\emph{Artificial Force-fields:}
Much research has focused on generating non-uniform artificial force-fields that can be used to rearrange passive components.  Applications have included techniques to design shear forces for sensorless manipulation of a single object by~\citet{lamiraux+2001:ra}.  
\citet{Vose2009a,vose2012sliding} demonstrated a collection of 2D force fields generated by 6DOF vibration inputs to a rigid plate.  These force fields, including shear forces, could be used as a set of primitives for motion control to steer the formation of multiple objects, but required multi-modal, position-dependent control.
\section{Conclusion and Future Work}\label{sec:conclusion}

This paper presented techniques for controlling the shape of a swarm of robots using global inputs and interaction with boundary friction forces.  
The paper provided algorithms for precise position control, as well as demonstrations of efficient covariance control. 
Extending algorithms \ref{alg:XControl} and \ref{alg:PosControl2Robots} to 3D is straightforward but increases the complexity.
Future efforts should be directed toward improving the technology and tailoring it to specific robot applications.

  With regard to technological advances, this includes designing controllers that efficiently regulate $\sigma_{xy}$, perhaps using Lyapunov-inspired controllers as in \citet{kim2015imparting}. 
 Additionally, this paper assumed that wall friction was nearly infinite.  The algorithms require retooling to handle small $\mu_f$ friction coefficients.  It may be possible to rank controllability as a function of friction.
  In hardware, the wall friction can be varied by laser-cutting boundary walls with different of profiles.

\section*{Acknowledgments}
This work was supported by the National Science Foundation under Grant No.\ \href{http://nsf.gov/awardsearch/showAward?AWD_ID=1553063}{ [IIS-1553063]}.
{\footnotesize
\bibliographystyle{plainnat}
\bibliography{IEEEabrv,ShapingSwarmFrictionSharedInput}
}

\title{\huge{ \emph{Supplement to} 
Algorithms For Shaping a Particle Swarm\\ With a Shared Control Input Using Boundary Interaction}}

\author{Shiva Shahrokhi, Arun Mahadev, and Aaron T. Becker}

\newpage
\maketitle

{\huge{ \emph{Supplement to} 
Algorithms For Shaping a Particle Swarm\\ With a Shared Control Input Using Boundary Interaction}}\\

\begin{abstract}
Includes algorithms and equations too lengthy for main paper, but potentially useful for the community.
Also links to videos and demonstration code for the algorithms.

\end{abstract}

\IEEEpeerreviewmaketitle

\section{Introduction}
This supplement gives overviews of the videos and code in 
\S \ref{sec:Videos}, 
provides the algorithm for $y$ position control of two robots in
\S \ref{sec:2robotWallFriction},
and gives gull analytical models for fluid settling in square-shaped tanks in
\S \ref{sec:fluidInPlanarRegion}.

\section{Supplementary Videos}\label{sec:Videos}
Five videos animate the key algorithms in this paper.

\subsection{Robot Swarm in a Circle under Gravity}
The video \emph{Robot Swarm in a Circle under Gravity} shows the stable configuration of a swarm under a constant global input.  Animated plots show mean, variance, covariance, and correlation for a swarm in a circular workspace.
Full resolution video: \url{https://youtu.be/nPFAjVIOxYc}.
An online demonstration and source code of the algorithm are at \citet{Zhao2016mathematicaSquare}.

\subsection{Distribution of Robot Swarm in Square under Gravity }
The video \emph{Distribution of Robot Swarm in Square under Gravity } shows the stable configuration of a swarm under a constant global input.  Animated plots show mean, variance, covariance, and correlation for a swarm in a square workspace.
Full resolution video: \url{https://youtu.be/ZEksDxLpAzg}.
An online demonstration and source code of the algorithm are at \citet{Zhao2016mathematica}.

\subsection{Steering 2 Particles with Shared Controls Using Wall Friction}
Animates Algs. 1, 2, 3 in Mathematica to show how two robots can be arbitrarily positioned in a square workspace. In this video the desired initial and ending positions of the two robots are manipulated, and the path that the robots should follow is drawn. The video ends with an extreme case where the robots must exchange positions. 
Full resolution video: \url{https://youtu.be/5TWlw7vThsM}.
An online demonstration and source code of the algorithm are at \citet{Shahrokhi2015mathematicaParticle}.

\subsection{Arranging a robot swarm with global inputs and wall friction [discrete] }
\begin{figure}
\begin{center}
	\includegraphics[width=1.0\columnwidth]{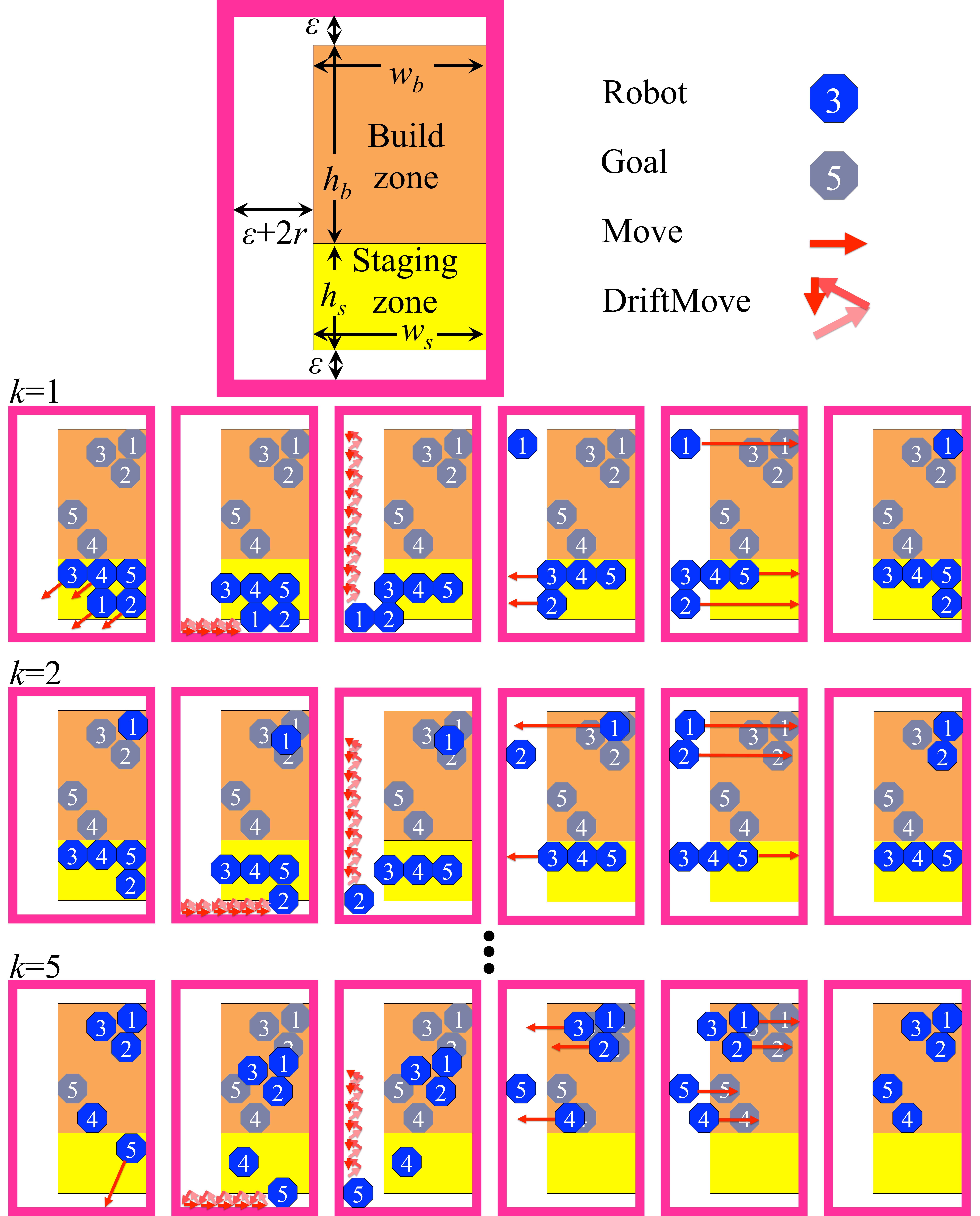}
\end{center}
\vspace{-1em}
\caption{\label{fig:PositionNrobotsCont}
Illustration of Alg.\ \ref{alg:PosControlNRobots}, $n$ robot position control  using wall friction.
}
\end{figure}

An implementation of Alg. 4  in {\sc Matlab} that illustrates how the two robots positioning algorithm is extendable to $n$ robots. In this video all  robots gets the same input, but by exploiting wall friction each robot reaches its goal, the formation "UH".
Full resolution video: \url{https://youtu.be/uhpsAyPwKeI}.
Full code is available at \citet{Arun2015}.
Note that this code uses discretized version of Algorithm 3.  The continuous-movement version is illustrated in Fig.\ref{fig:PositionNrobotsCont}.

\subsection{AutomaticCovControl.mp4}
A closed-loop controller that steers a swarm of particles to a desired covariance,  implemented with a box2D simulator. In this video the green ellipse is the desired covariance ellipse, the red ellipse is the current covariance ellipse of the swarm and the red dot is the mean position of the robots. Robots follow the algorithm to achieve the desired values for $\sigma_{goalxy}$, $\sigma_x^2$ and $\sigma_y^2$.

\section{ Algorithm for generating desired $y$ spacing between two robots using wall friction}\label{sec:2robotWallFriction}
\begin{algorithm}
\caption{GenerateDesired$y$-spacing($s_1,s_2,e_1,e_2,L$)}\label{alg:YControl}
\begin{algorithmic}[1]
\Require Knowledge of starting $(s_1,s_2)$ and ending $(e_1,e_2)$ positions of  two robots. 
$(0,0)$ is bottom corner, $s_1$ is rightmost robot, 
 $L$ is length of the walls. Current position of the robots are $(r_1,r_2)$.
\Ensure   $ r_{1x} - r_{2x}  \equiv s_{1x} - s_{2x} $   
\State $ \Delta s_y  \gets s_{1y} - s_{2y} $
\State $ \Delta e_y \gets e_{1y} - e_{2y} $
\State $ r_1 \gets s_1$, $ r_2 \gets s_2$
\If {$\Delta e_y < 0 $ }
\State $ m \gets ( L-\max( r_{1y},r_{2y}) ,0)   $ \Comment{Move to top wall}
\Else 
\State  $ m \gets ( -\min( r_{1y},r_{2y}),0 )    $ \Comment{Move to bottom wall}
\EndIf
\State $m  \gets  m + (0, -\min( r_{1x},r_{2x} ))$ \Comment{Move to left}
\State $ r_1 \gets r_1+m$, $ r_2 \gets r_2+m$ \Comment{Apply move}
\If {$\Delta e_y - (r_{1y} - r_{2y} ) > 0 $}
\State $ m \gets (\min(|\Delta e_y - \Delta s_y |, L- r_{1y}), 0)$  \Comment{Move top}
\Else
\State $ m \gets (-\min(|\Delta e_y - \Delta s_y |, r_{1y}), 0)$\Comment{Move bottom}
\EndIf 
\State $m  \gets  m + (0, \epsilon)$ \Comment{Move right}
\State $ r_1 \gets r_1+m$, $ r_2 \gets r_2+m$ \Comment{Apply move}
\State $\Delta r_y = r_{1y} - r_{2y}$
\If {$\Delta r_y \equiv \Delta e_y$} 
\State   $ m \gets (e_{1x}-r_{1x}, e_{1y}-r_{1y})$
\State $ r_1 \gets r_1+m$, $ r_2 \gets r_2+m$ \Comment{Apply move}
\State  \Return $(r_1,r_2)$
\Else   
\State \Return GenerateDesired$y$-spacing($r_1,r_2,e_1,e_2,L$)
\EndIf
\end{algorithmic}
\end{algorithm}

\section{Calculations for modeling swarm as fluid in a simple planar workspace}\label{sec:fluidInPlanarRegion}
Two workspaces are used, a square and a circular workspace.

\subsection{Square Workspace}
This section provides formulas for the mean, variance,  covariance and correlation of a very large swarm of robots as they move inside a square workplace under the influence of gravity pointing in the direction $\beta$. The swarm is large, but the robots are small in comparison, and together cover an area of constant volume $A$. Under a global input such as gravity, they flow like water, moving to a side of the workplace and forming a polygonal shape. The workspace is 

The range of possible angles for the global input angle $\beta $ is [0,2$\pi $). In this range of angles, the swarm assumes eight different polygonal shapes. The shapes alternate between triangles and trapezoids when the area $A$$<$1/2, and alternate between squares with one corner removed and trapezoids when $A$$>$1/2.

Two representative formulas are attached, the outline of the swarm shapes in \eqref{tab:SquareRobotRegions} and $\bar{x}(\beta,A)$ in \eqref{tab:SquareXMean}.

\begin{figure}[h]
\begin{center}
\includegraphics[width=\columnwidth]{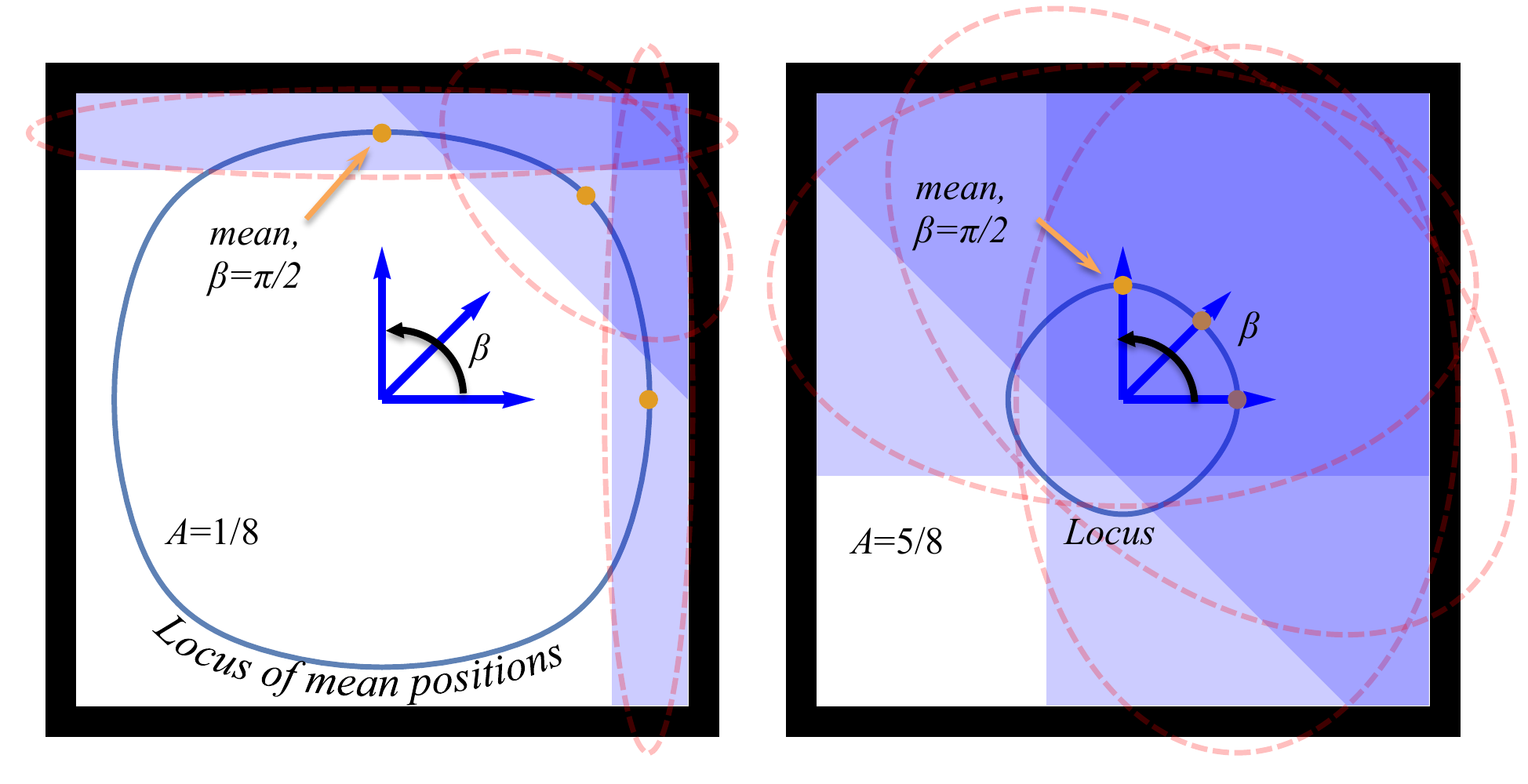} 
\caption{A swarm in a square workspace under a constant global input assumes either a triangular or a trapezoidal shape if $A<1/2$.  If $A>1/2$ the swarm is either a squares with one corner removed or a trapezoidal  shape.}
\label{fig:friction}
\end{center}
\end{figure} 

\begin{table*}
\begin{align}
\bar{x}(\beta,A) = A\leq \frac{1}{2}: &\begin{cases}
 -\frac{\tan ^2(\beta )}{24 A}-\frac{A}{2}+1 & 0\leq \beta \leq \tan ^{-1}(2 A)\lor 2 \pi -\tan ^{-1}(2 A)<\beta \leq 2 \pi  \\
 1-\frac{1}{3} \sqrt{2} \sqrt{A \tan (\beta )} & \tan ^{-1}(2 A)<\beta \leq \frac{\pi }{2}-\tan ^{-1}(2 A) \\
 \frac{\cot (\beta )}{12 A}+\frac{1}{2} & \frac{\pi }{2}-\tan ^{-1}(2 A)<\beta \leq \tan ^{-1}(2 A)+\frac{\pi }{2} \\
 \frac{1}{3} \sqrt{2} \sqrt{-A \tan (\beta )} & \tan ^{-1}(2 A)+\frac{\pi }{2}<\beta \leq \pi -\tan ^{-1}(2 A) \\
 \frac{\tan ^2(\beta )}{24 A}+\frac{A}{2} & \pi -\tan ^{-1}(2 A)<\beta \leq \tan ^{-1}(2 A)+\pi  \\
 \frac{1}{3} \sqrt{2} \sqrt{A \tan (\beta )} & \tan ^{-1}(2 A)+\pi <\beta \leq \frac{3 \pi }{2}-\tan ^{-1}(2 A) \\
 \frac{1}{2}-\frac{\cot (\beta )}{12 A} & \frac{3 \pi }{2}-\tan ^{-1}(2 A)<\beta \leq \tan ^{-1}(2 A)+\frac{3 \pi }{2} \\
 1-\frac{1}{3} \sqrt{2} \sqrt{-A \tan (\beta )} & \tan ^{-1}(2 A)+\frac{3 \pi }{2}<\beta \leq 2 \pi -\tan ^{-1}(2 A) \\
\end{cases} \nonumber\\
\frac{1}{2}<A<1:&\begin{cases}
 -\frac{\tan ^2(\beta )}{24 A}-\frac{A}{2}+1 & 0\leq \beta \leq \tan ^{-1}\left(\frac{1}{2},1-A\right)\lor 2 \pi -\tan ^{-1}\left(\frac{1}{2},1-A\right)<\beta \leq 2 \pi  \\
 \frac{2 \sqrt{2} \sqrt{(1-A) \tan (\beta )} (A-1)+3}{6 A} & \tan ^{-1}\left(\frac{1}{2},1-A\right)<\beta \leq \frac{\pi }{2}-\tan ^{-1}\left(\frac{1}{2},1-A\right) \\
 \frac{6 A+\cot (\beta )}{12 A} & \frac{\pi }{2}-\tan ^{-1}\left(\frac{1}{2},1-A\right)<\beta \leq \tan ^{-1}\left(\frac{1}{2},1-A\right)+\frac{\pi }{2} \\
 \frac{-2 \sqrt{2} \sqrt{(A-1) \tan (\beta )} (A-1)+6 A-3}{6 A} & \tan ^{-1}\left(\frac{1}{2},1-A\right)+\frac{\pi }{2}<\beta \leq \pi -\tan ^{-1}\left(\frac{1}{2},1-A\right) \\
 \frac{\tan ^2(\beta )}{24 A}+\frac{A}{2} & \pi -\tan ^{-1}\left(\frac{1}{2},1-A\right)<\beta \leq \tan ^{-1}\left(\frac{1}{2},1-A\right)+\pi  \\
 \frac{2 \sqrt{2} \sqrt{(1-A) \tan (\beta )} (1-A)+6 A-3}{6 A} & \tan ^{-1}\left(\frac{1}{2},1-A\right)+\pi <\beta \leq \frac{3 \pi }{2}-\tan ^{-1}\left(\frac{1}{2},1-A\right) \\
 \frac{1}{2}-\frac{\cot (\beta )}{12 A} & \frac{3 \pi }{2}-\tan ^{-1}\left(\frac{1}{2},1-A\right)<\beta \leq \tan ^{-1}\left(\frac{1}{2},1-A\right)+\frac{3 \pi }{2} \\
 \frac{2 \sqrt{2} \sqrt{(A-1) \tan (\beta )} (A-1)+3}{6 A} & \tan ^{-1}\left(\frac{1}{2},1-A\right)+\frac{3 \pi }{2}<\beta \leq 2 \pi -\tan ^{-1}\left(\frac{1}{2},1-A\right) \\
\end{cases}
 \nonumber \\
A=1: &\frac{1}{2}
\end{align}
\protect\caption{$\bar{x}$ in a unit-square workspace}
\label{tab:SquareXMean}
\end{table*}

\begin{table*}
\tiny
\begin{align}
\text{RobotRegion}(\beta,A)= \nonumber 
A\leq \frac{1}{2}:&
\begin{cases}
 \left(
\begin{array}{cc}
 1 & 0 \\
 1 & 1 \\
 -A-\frac{\tan (\beta )}{2}+1 & 1 \\
 -A+\frac{\tan (\beta )}{2}+1 & 0 \\
\end{array}
\right) & 0\leq \beta \leq \tan ^{-1}(2 A)\lor 2 \pi -\tan ^{-1}(2 A)<\beta \leq 2 \pi  \\
 \left(
\begin{array}{cc}
 1 & 1 \\
 1-\sqrt{2} \sqrt{A \tan (\beta )} & 1 \\
 1 & 1-\sqrt{2} \sqrt{A \cot (\beta )} \\
\end{array}
\right) & \tan ^{-1}(2 A)<\beta \leq \frac{\pi }{2}-\tan ^{-1}(2 A) \\
 \left(
\begin{array}{cc}
 1 & 1 \\
 0 & 1 \\
 0 & -A+\frac{\cot (\beta )}{2}+1 \\
 1 & -A-\frac{\cot (\beta )}{2}+1 \\
\end{array}
\right) & \frac{\pi }{2}-\tan ^{-1}(2 A)<\beta \leq \tan ^{-1}(2 A)+\frac{\pi }{2} \\
 \left(
\begin{array}{cc}
 0 & 1 \\
 \sqrt{2} \sqrt{-A \tan (\beta )} & 1 \\
 0 & 1-\sqrt{2} \sqrt{-A \cot (\beta )} \\
\end{array}
\right) & \tan ^{-1}(2 A)+\frac{\pi }{2}<\beta \leq \pi -\tan ^{-1}(2 A) \\
 \left(
\begin{array}{cc}
 0 & 0 \\
 0 & 1 \\
 A-\frac{\tan (\beta )}{2} & 1 \\
 A+\frac{\tan (\beta )}{2} & 0 \\
\end{array}
\right) & \pi -\tan ^{-1}(2 A)<\beta \leq \tan ^{-1}(2 A)+\pi  \\
 \left(
\begin{array}{cc}
 0 & 0 \\
 0 & \sqrt{2} \sqrt{A \cot (\beta )} \\
 \sqrt{2} \sqrt{A \tan (\beta )} & 0 \\
\end{array}
\right) & \tan ^{-1}(2 A)+\pi <\beta \leq \frac{3 \pi }{2}-\tan ^{-1}(2 A) \\
 \left(
\begin{array}{cc}
 0 & 0 \\
 1 & 0 \\
 1 & A-\frac{\cot (\beta )}{2} \\
 0 & A+\frac{\cot (\beta )}{2} \\
\end{array}
\right) & \frac{3 \pi }{2}-\tan ^{-1}(2 A)<\beta \leq \tan ^{-1}(2 A)+\frac{3 \pi }{2} \\
 \left(
\begin{array}{cc}
 1 & 0 \\
 1-\sqrt{2} \sqrt{-A \tan (\beta )} & 0 \\
 1 & \sqrt{2} \sqrt{-A \cot (\beta )} \\
\end{array}
\right) & \tan ^{-1}(2 A)+\frac{3 \pi }{2}<\beta \leq 2 \pi -\tan ^{-1}(2 A) \\
\end{cases}
\nonumber \\
\frac{1}{2}<A<1:&
\begin{cases}
 \left(
\begin{array}{cc}
 1 & 0 \\
 1 & 1 \\
 (1-A)-\frac{\tan (\beta )}{2} & 1 \\
 (1-A)+\frac{\tan (\beta )}{2} & 0 \\
\end{array}
\right) & 0\leq \beta \leq \tan ^{-1}\left(\frac{1}{2},1-A\right)\lor 2 \pi -\tan ^{-1}\left(\frac{1}{2},1-A\right)<\beta \leq 2 \pi  \\
 \left(
\begin{array}{cc}
 1 & 0 \\
 1 & 1 \\
 0 & 1 \\
 0 & \sqrt{2} \sqrt{(1-A) \cot (\beta )} \\
 \sqrt{2} \sqrt{(1-A) \tan (\beta )} & 0 \\
\end{array}
\right) & \tan ^{-1}\left(\frac{1}{2},1-A\right)<\beta \leq \frac{\pi }{2}-\tan ^{-1}\left(\frac{1}{2},1-A\right) \\
 \left(
\begin{array}{cc}
 0 & 1 \\
 1 & 1 \\
 1 & (1-A)-\frac{\cot (\beta )}{2} \\
 0 & (1-A)+\frac{\cot (\beta )}{2} \\
\end{array}
\right) & \frac{\pi }{2}-\tan ^{-1}\left(\frac{1}{2},1-A\right)<\beta \leq \tan ^{-1}\left(\frac{1}{2},1-A\right)+\frac{\pi }{2} \\
 \left(
\begin{array}{cc}
 1 & 1 \\
 0 & 1 \\
 0 & 0 \\
 1-\sqrt{2} \sqrt{-(1-A) \tan (\beta )} & 0 \\
 1 & \sqrt{2} \sqrt{-(1-A) \cot (\beta )} \\
\end{array}
\right) & \tan ^{-1}\left(\frac{1}{2},1-A\right)+\frac{\pi }{2}<\beta \leq \pi -\tan ^{-1}\left(\frac{1}{2},1-A\right) \\
 \left(
\begin{array}{cc}
 0 & 0 \\
 0 & 1 \\
 -(1-A)-\frac{\tan (\beta )}{2}+1 & 1 \\
 -(1-A)+\frac{\tan (\beta )}{2}+1 & 0 \\
\end{array}
\right) & \pi -\tan ^{-1}\left(\frac{1}{2},1-A\right)<\beta \leq \tan ^{-1}\left(\frac{1}{2},1-A\right)+\pi  \\
 \left(
\begin{array}{cc}
 1 & 0 \\
 0 & 0 \\
 0 & 1 \\
 1-\sqrt{2} \sqrt{(1-A) \tan (\beta )} & 1 \\
 1 & 1-\sqrt{2} \sqrt{(1-A) \cot (\beta )} \\
\end{array}
\right) & \tan ^{-1}\left(\frac{1}{2},1-A\right)+\pi <\beta \leq \frac{3 \pi }{2}-\tan ^{-1}\left(\frac{1}{2},1-A\right) \\
 \left(
\begin{array}{cc}
 1 & 0 \\
 0 & 0 \\
 0 & -(1-A)+\frac{\cot (\beta )}{2}+1 \\
 1 & -(1-A)-\frac{\cot (\beta )}{2}+1 \\
\end{array}
\right) & \frac{3 \pi }{2}-\tan ^{-1}\left(\frac{1}{2},1-A\right)<\beta \leq \tan ^{-1}\left(\frac{1}{2},1-A\right)+\frac{3 \pi }{2} \\
 \left(
\begin{array}{cc}
 0 & 0 \\
 1 & 0 \\
 1 & 1 \\
 \sqrt{2} \sqrt{-(1-A) \tan (\beta )} & 1 \\
 0 & 1-\sqrt{2} \sqrt{-(1-A) \cot (\beta )} \\
\end{array}
\right) & \tan ^{-1}\left(\frac{1}{2},1-A\right)+\frac{3 \pi }{2}<\beta \leq 2 \pi -\tan ^{-1}\left(\frac{1}{2},1-A\right) \\
\end{cases},\nonumber\\
A=1:&\left(
\begin{array}{cc}
 1 & 0 \\
 0 & 0 \\
 0 & 1 \\
 1 & 1 \\
\end{array}
\right)
\end{align}
\protect\caption{RobotRegions in a unit-square workspace}
\label{tab:SquareRobotRegions}
\end{table*}

\subsection{Circle Workspace}
The area under a chord of a circle is the area of a sector less the area of the triangle originating at the circle center: 
$A=S(sector)-S(triangle)=1/2 LR-1/2 C(1-h)$, thus
\begin{align}
A=(1/2)\left[LR-c(R-h)\right]
\end{align}
where $L$ is arc length, $c$ is chord length, $R$ is radius and $h$ is height. Solving for $L$ and $C$ gives
\begin{align}
L&=2 \cos ^{-1}(1-h)\\
C&=2\sqrt{h(2-h)}
\end{align}
Therefore the area under a chord is
\begin{align}
\cos ^{-1}(1-h)-(1-h) \sqrt{(2-h) h}
\end{align}

For a circular workspace, with $\beta = 0$, the variance of $x$ and $y$ are:
{\tiny
\begin{align}
&\sigma_x^2(h)=\frac{64 (h-2)^3 h^3}{144 \left(\sqrt{-(h-2) h} (h-1)+\arccos(1-h)\right)^2} +\nonumber\\
&\frac{9 \left(\sqrt{-(h-2) h} (h-1)+\arccos(1-h)\right) \left(\sin \left(4 \arcsin(1-h)\right)+4 \arccos(1-h)\right)}{144 \left(\sqrt{-(h-2) h} (h-1)+\arccos(1-h)\right)^2}
\end{align}}

{\tiny
\begin{align}
\sigma_y^2(h)=
\frac{12 \arccos(1-h)-8 \sin \left(2 \arccos(1-h)\right)+\sin \left(4 \arccos(1-h)\right)}{48 \left(\sqrt{-(h-2) h} (h-1)+\arccos(1-h)\right)}
\end{align}}

For $\beta = 0$, $\sigma_{xy}=0$. These values can be rotated to calculate $\sigma_x^2(\beta,h),\sigma_y^2(\beta,h),$ and $\sigma_{xy}(\beta,h)$.

\section*{Acknowledgments}
This work was supported by the National Science Foundation under Grant No.\ \href{http://nsf.gov/awardsearch/showAward?AWD_ID=1553063}{ [IIS-1553063]}.

\bibliographystyle{plainnat}
\footnotesize
\bibliography{IEEEabrv,ShapingSwarmFrictionSharedInput}

\end{document}